\documentclass[dvipsnames]{fairmeta}

\usepackage{hyperref}
\usepackage{makecell}
\usepackage{siunitx}
\usepackage{framed}
\usepackage{pifont}
\usepackage{graphicx,subcaption}
\usepackage{svg}
\usepackage{xspace}
\usepackage{nicematrix}
\usepackage{nicefrac}
\usepackage{wrapfig}
\usepackage{enumitem}
\usepackage{amssymb}
\usepackage{tabularx}
\usepackage{caption}

\newsavebox{\ghlogo}
\savebox{\ghlogo}{\includegraphics[height=0.95em]{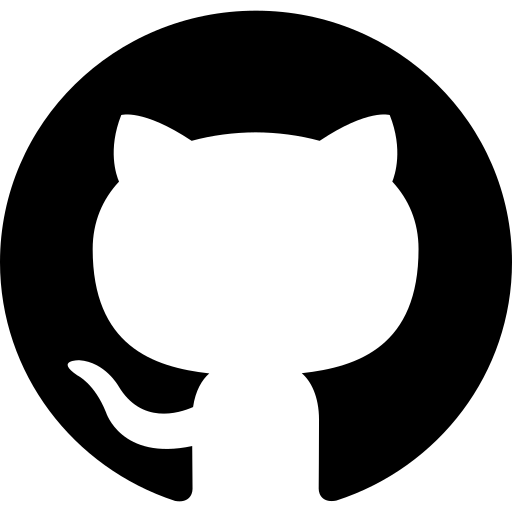}}
\newsavebox{\hflogo}
\savebox{\hflogo}{\includegraphics[height=1.2em]{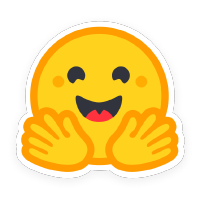}}

\DeclareRobustCommand{\ghicon}{\raisebox{-0.25em}{\makebox[1.5em][c]{\usebox{\ghlogo}}}}
\DeclareRobustCommand{\hficon}{\raisebox{-0.25em}{\makebox[1.5em][c]{\usebox{\hflogo}}}}

\title{POINTS-GUI-G: GUI-Grounding Journey}

\author[1]{Zhongyin Zhao$^\ast$}

\author[1]{Yuan Liu$^{\ast\dagger}$}

\author[1]{Yikun Liu}

\author[1]{Haicheng Wang}

\author[1]{Le Tian}

\author[1]{Xiao Zhou}

\author[1]{Yangxiu You}

\author[1]{Zilin Yu}

\author[1]{Yang Yu}

\author[1]{Jie Zhou}

\affiliation[1]{WeChat AI}

\contribution[\ast]{Equal Contribution}
\contribution[\dagger]{Corresponding Author}
\renewcommand\contribution[2][]{\addtolist[#1]{#2}{\contributionlist}{\contributionformat}{\\[1mm]}}
\contribution{\ghicon\ Github: \url{https://github.com/Tencent/POINTS-GUI}}
\contribution{\hficon\ HuggingFace: \url{https://huggingface.co/tencent/POINTS-GUI-G}}

\abstract{
The rapid advancement of vision-language models has catalyzed the emergence of GUI agents, which hold immense potential for automating complex tasks—from online shopping to flight booking—thereby alleviating the burden of repetitive digital workflows. As a foundational capability, GUI grounding is typically established as a prerequisite for end-to-end task execution. It enables models to precisely locate interface elements, such as text and icons, to perform accurate operations like clicking and typing. Unlike prior works that fine-tune models already possessing strong spatial awareness (e.g., Qwen3-VL), we aim to master the full technical pipeline by starting from a base model with minimal grounding ability, such as POINTS-1.5. We introduce \textsc{POINTS-GUI-G-8B}, which achieves state-of-the-art performance with scores of 59.9 on ScreenSpot-Pro, 66.0 on OSWorld-G, 95.7 on ScreenSpot-v2, and 49.9 on UI-Vision. Our model's success is driven by three key factors: (1) \textbf{Refined Data Engineering}, involving the unification of diverse open-source datasets format alongside sophisticated strategies for augmentation, filtering, and difficulty grading; (2) \textbf{Improved Training Strategies}, including continuous fine-tuning of the vision encoder to enhance perceptual accuracy and maintaining resolution consistency between training and inference; and (3) \textbf{Reinforcement Learning (RL) with Verifiable Rewards}. While RL is traditionally used to bolster reasoning, we demonstrate that it significantly improves precision in the perception-intensive GUI grounding task. Furthermore, GUI grounding provides a natural advantage for RL, as rewards are easily verifiable and highly accurate.
}

\date{Feb 6, 2026}
\correspondence{\email{bensenliu@tencent.com}}

\begin{document}
\maketitle
\section{Introduction}
Driven by rapid advancements in Vision-Language Models (VLMs)~\citep{liu2024points, liu2024points1, liu2024rethinking, yang2025qwen3, zhu2025internvl3, wang2025internvl3_5, zeng2025glm, Qwen2.5-VL, team2025kimi, li2024llava, liu2024llavanext}, Graphical User Interface (GUI) agents have emerged as a transformative research frontier. Deploying these agents in real-world production environments enables the automation of complex tasks—ranging from ticket booking to online shopping—thereby streamlining daily workflows. Compared to MCP-based agents, GUI agents~\citep{qin2025ui, wang2025ui, ye2025mobile, zhou2025maiuitechnicalreportrealworld, gu2025ui, yan2025step} offer broader applicability as they interact directly with interfaces without requiring modifications to existing internet infrastructure. To execute tasks effectively, an agent must decompose high-level goals into discrete steps, each requiring the precise localization of interface elements like buttons and input fields. Consequently, GUI grounding is a fundamental prerequisite. While current research often leverages models with pre-existing, well-optimized grounding capabilities (e.g., Qwen3-VL) to focus on end-to-end execution, this approach may overlook critical technical insights essential for iterative development. To achieve a comprehensive, ``full-stack'' mastery of GUI agent development, we deliberately start with POINTS-1.5, a model lacking native grounding capabilities. We build the GUI grounding capability for the WePOINTS series from the ground up through three pillars: 1) refined data engineering, 2) precise training strategies, and 3) reinforcement learning (RL) with verifiable rewards.

\begin{figure}[!htbp]
\centering
\includegraphics[width=\linewidth,scale=1.00]{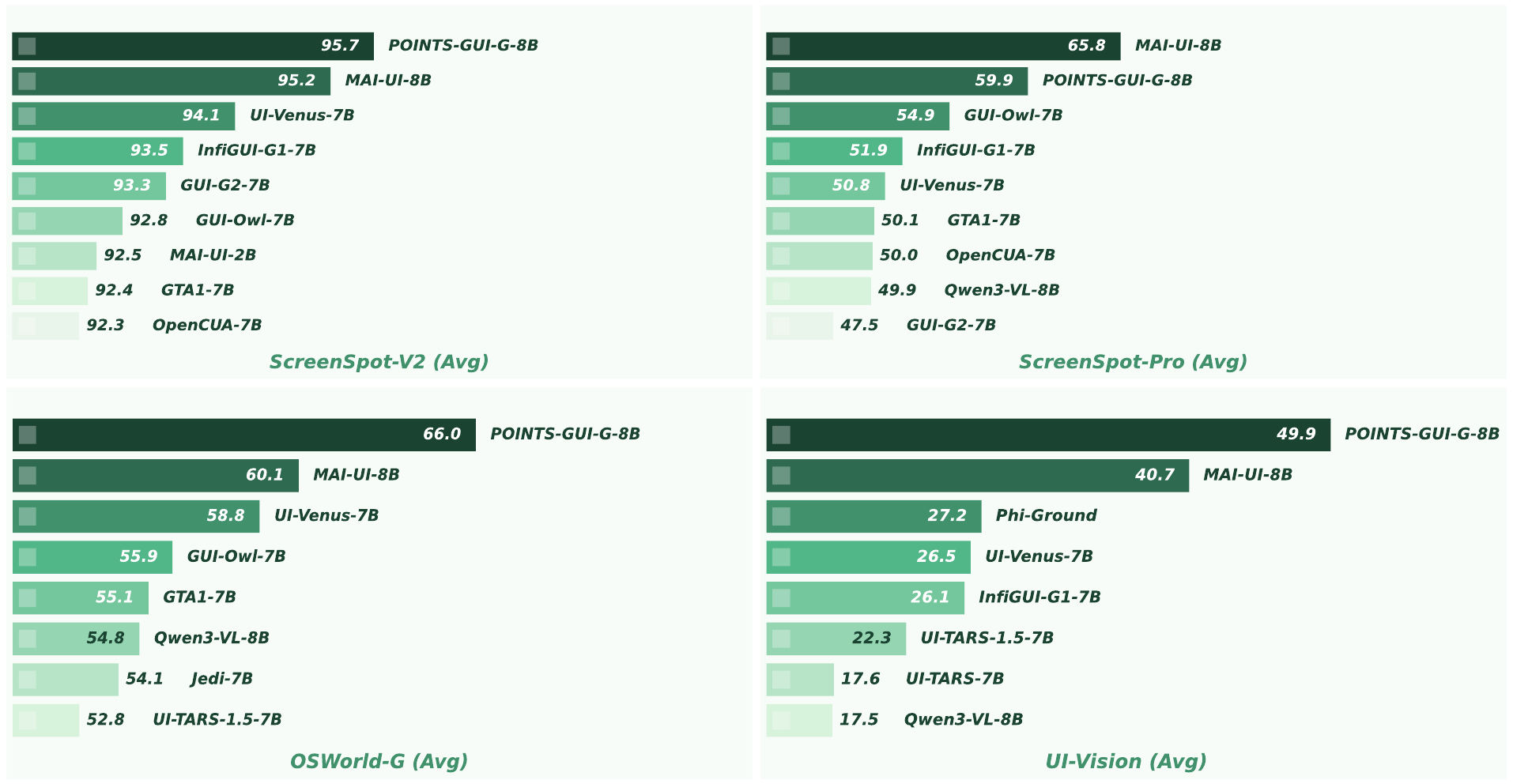}
\caption{\textbf{Comparison with existing models of comparable size on ScreenSpot-v2~\citep{wu2024atlas}, SreenSpot-Pro~\citep{li2025screenspotpro}, MMBench-GUI-L2~\citep{Wang2025MMBenchGUIHM}, and OSWorld-G~\citep{xie2025scalingcomputerusegroundinguser}.}}
\label{fig:fig1}
\end{figure}

\paragraph{\textbf{Refined Data Engineering}} We implement a multi-stage data engineering pipeline to address the limitations of existing GUI grounding datasets. 1) \textbf{Standardization}: While numerous open-source datasets \citep{feizi2025groundingcomputeruseagents, mu2025gui, chen2025guicourse, xie2025scalingcomputerusegroundinguser, liu2024harnessingwebpageuistextrich, wu2024atlas, cheng2024seeclick, bai2021uibert, wu2023webui} enhance agent grounding, they exhibit significant heterogeneity in coordinate scales (e.g., normalized vs. raw pixels) and task formats. We unify these by normalizing all coordinates to a $[0, 1]$ range and reformatting diverse tasks into a consistent UI element localization task, thereby mitigating training interference. 2) \textbf{Noise Reduction}: Our analysis reveals that existing datasets—whether constructed via URL crawling (e.g., from FineWeb \citep{penedo2024the}) or model-based annotation of screenshots (e.g., RICO)—often contain significant noise. To ensure data quality, we developed an automated filtering solution to prune inaccurate annotations. 3) \textbf{Complexity Enhancement}: To push the model's performance ceiling, we increase training difficulty by filtering out ``simple'' samples—those with sparse layouts or oversized clickable regions—using metrics such as element layout entropy and occupancy ratios. Furthermore, we synthesize challenging data by: (i) rendering professional software with complex, small-scale elements, and (ii) layering multiple pages onto a desktop-style interface to introduce realistic distractors.

\paragraph{\textbf{Improved Training Strategies}} In previous iterations of the WePOINTS series~\citep{liu2024points1, liu-etal-2025-points}, the vision encoder was kept frozen during training. However, GUI grounding is a perception-heavy task where the quality of extracted image features directly determines model performance. Our experiments demonstrate that general-purpose vision encoders, such as Qwen2VL-ViT~\citep{wang2024qwen2}, are not fully optimized for this specialized domain. Consequently, we fully unfreeze the vision encoder during training, leading to substantial performance gains. Additionally, we address the often-overlooked factor of image resolution. Previously, in order to accelerate training and save memory, we restrict the maximum image resolution below $2000 \times 2000$ pixels. As a result, models trained on these data often suffer from significant performance degradation when encountering higher-resolution images, such as those in ScreenSpot-Pro~\citep{li2025screenspotpro}, during inference. To mitigate this, we propose two strategies: (1) increasing the maximum training resolution cap to 
$3072 \times 3072$ and (2) constraining the inference resolution to under $2000 \times 2000$ pixels. Together, these strategies yielded a performance boost of over 10 points on the ScreenSpot-Pro benchmark.

\paragraph{\textbf{Reinforcement Learning with Verifiable Rewards}} Reinforcement Learning with Verifiable Rewards (RLVR) has been primarily utilized to enhance the reasoning and agentic capabilities of large-scale models \citep{shao2024deepseekmath, meng2025mm, zeng2025glm, chen2025minimax}. Compared to domains such as mathematical theorem proving or open-ended generation, GUI grounding offers distinct advantages for reinforcement learning. The output space is highly constrained—consisting of points or bounding boxes—enabling the definition of precise, objective reward functions. Diverging from certain prior approaches \citep{chen2025ui} and following GTA1 \citep{yang2025gta1}, we do not require the model to generate a reasoning trace prior to outputting the target coordinates. Although GUI grounding is fundamentally a perception-based task, we observe that RLVR yields consistent and substantial improvements over the supervised baseline. Furthermore, we implement a curriculum-based strategy, progressively increasing the difficulty of training samples to stabilize and enhance the learning process.

In summary, we present \textsc{POINTS-GUI-G}, a new state-of-the-art model for GUI grounding. Our primary contributions are as follows:

\begin{enumerate}[label={\bf $\bullet$}, leftmargin=*, topsep=0.5ex, itemsep=0ex, partopsep=0.5ex, parsep=0.5ex, wide, labelindent=0pt]
    \item \textbf{Unified Data Engineering:} We reformat existing large-scale open-source GUI grounding datasets into a unified task formulation and annotation standard. Furthermore, we implement an automated pipeline for comprehensive data cleaning and introduce novel data construction strategies designed to enhance task complexity.
    \item \textbf{Resolution Optimization:} We revisit critical but previously overlooked factors in GUI grounding, specifically the image resolution mismatch between training and inference. We propose effective strategies to mitigate this discrepancy, resulting in substantial performance gains on high-resolution benchmarks such as ScreenSpot-Pro.
    \item \textbf{Reinforcement Learning:} We demonstrate the effectiveness of applying Reinforcement Learning (RL) with verifiable rewards to GUI grounding tasks, achieving consistent and significant improvements across benchmarks.
\end{enumerate}

We hope \textsc{POINTS-GUI-G} serves as a robust foundation for future GUI agent development and that our findings provide new perspectives on enhancing GUI grounding capabilities.
\section{Related Works}
\paragraph{\textbf{Vision-Language Models.}} Recent vision-language models (VLMs) have advanced rapidly across architectural efficiency and scaling~\citep{liu2024improved, liu2024points, liu2024points1, Bai2025Qwen3VLTR, wang2024qwen2, Qwen2.5-VL, zhu2025internvl3, wang2025internvl3_5}. Early milestones like BLIP-2~\citep{li2023blip} minimized training overhead by optimizing only the Q-Former, while LLaVA~\citep{liu2023visual} streamlined modality alignment through large-scale instruction tuning. Subsequent research has focused on visual fidelity; for instance, InternLM-XComposer2 and CogVLM~\citep{chen2024expanding, zhang2024internlm} utilize image tiling to handle higher resolutions. More recently, Qwen2-VL~\citep{wang2024qwen2} and Qwen2.5-VL~\citep{Qwen2.5-VL} implemented NaViT-style~\citep{dehghani2023patch} encoders to natively support arbitrary aspect ratios and resolutions. This progress has been further catalyzed by the development of comprehensive evaluation benchmarks like MMBench~\citep{liu2024mmbench}.

\paragraph{\textbf{GUI Grounding}} 
GUI grounding evaluates a model's ability to map natural language instructions to precise coordinates within a user interface. As a fundamental ability of GUI agents, it serves as a critical prerequisite for high-level task execution. Conventional approaches~\citep{wu2024atlas, qin2025ui, wang2025ui, Xu2024AguvisUP, xie2025scalingcomputerusegroundinguser, Gou2024NavigatingTD, yang-etal-2025-aria} primarily optimize this capability through extensive Supervised Fine-Tuning (SFT). However, following the success of Reinforcement Learning (RL)~\citep{shao2024deepseekmath, Schulman2017ProximalPO} in enhancing reasoning for domains such as mathematics and coding, several pioneering studies~\citep{Luo2025GUIR1A, wu2025gui, yang2025gta1} have begun integrating RL into the GUI grounding pipeline. These efforts have demonstrated substantial performance gains over SFT-only baselines—a finding consistent with the empirical observations made during the development of POINTS-GUI-G.
\section{Methods}

This section is organized into three parts. First, \autoref{sec:data_engineer} details our data engineering pipeline across three dimensions: (i) preprocessing operations applied to existing large-scale open-source GUI grounding datasets, (ii) filtering strategies used to curate a high-quality subset, and (iii) a complexity enhancement strategy designed to facilitate more effective learning in the later stages of training. Next, \autoref{sec:training_strategies} revisits critical but often overlooked aspects of the training process that significantly impact final model performance. Finally, \autoref{sec:rlvr} demonstrates how Reinforcement Learning with Verifiable Rewards (RLVR) further elevates the performance ceiling and describes the construction of the dataset used for this stage.

\subsection{Data Engineering}
\begin{figure}[!htbp]
\centering
\includegraphics[width=1.0\linewidth]{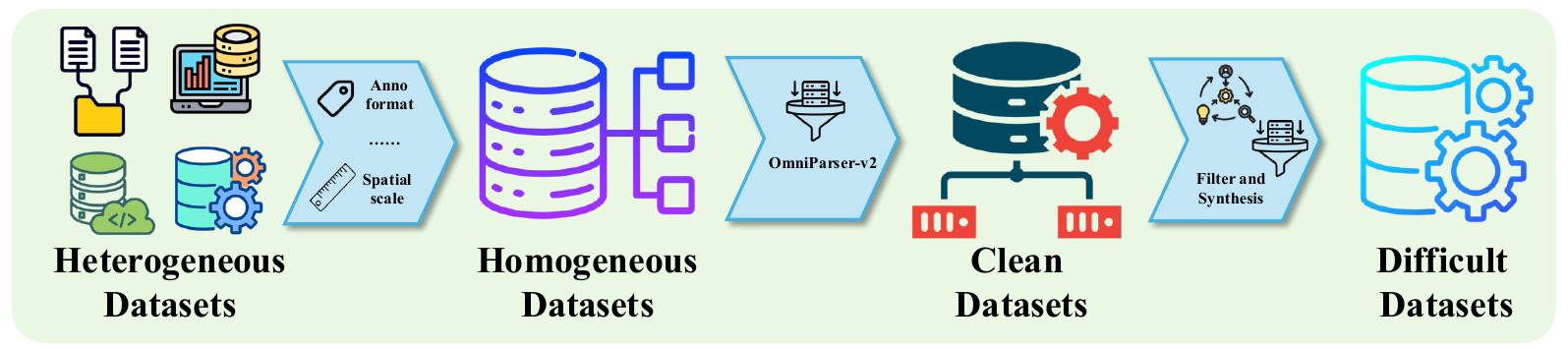}
\caption{\textbf{The three-stage data engineering pipeline:} Data preprocessing, data filtering, and complexity enhancement.}
\label{fig:fig2}
\end{figure}

\autoref{fig:fig2} illustrates the three-step data engineering pipeline.
\label{sec:data_engineer}

\begin{figure}[!htbp]
\centering
\includegraphics[width=\linewidth,scale=1.00]{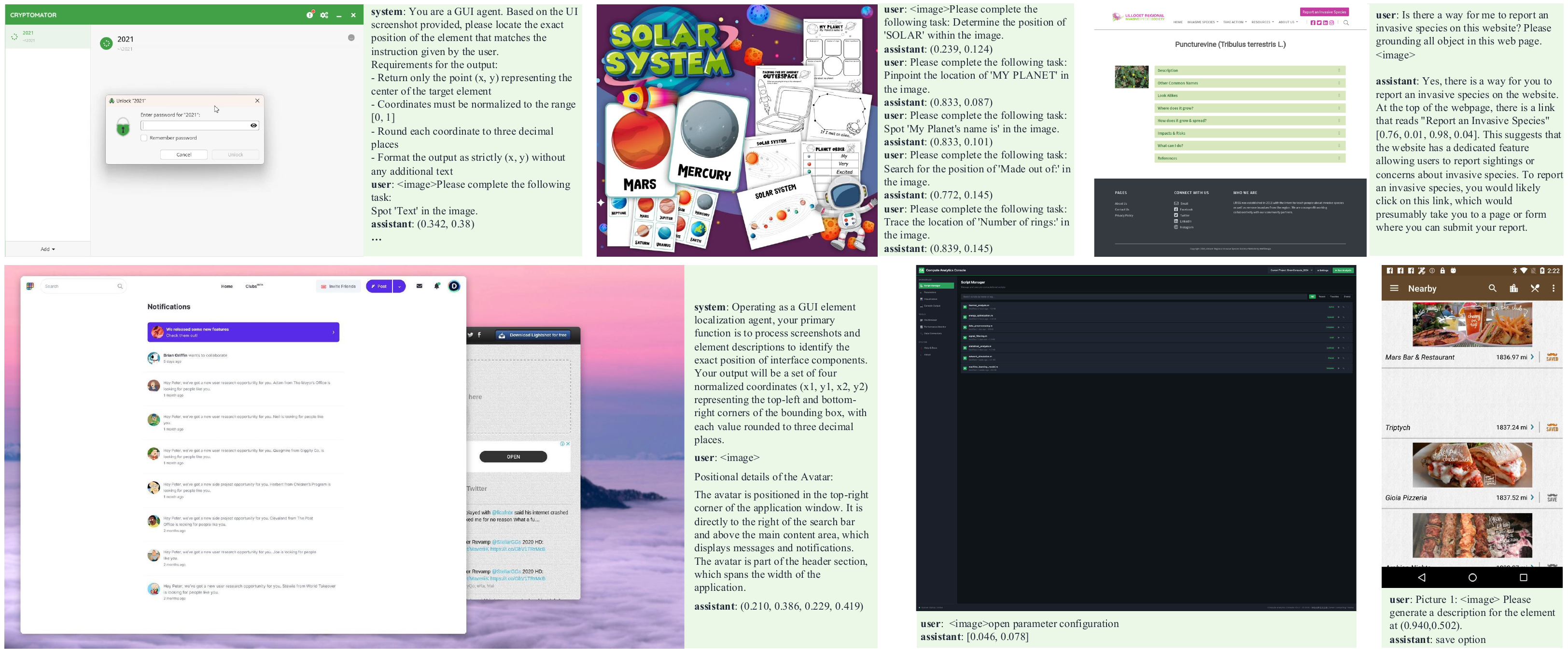}
\caption{Reformatted samples from existing open-source datasets. All spatial annotations are unified to a $[0, 1]$ scale.}
\label{fig:fig3}
\end{figure}

\paragraph{\textbf{Data Preprocessing}} Existing open-source GUI grounding datasets~\citep{wu2024atlas, wu2023webui, bai2021uibert, cheng2024seeclick, liu2024harnessingwebpageuistextrich} are highly heterogeneous, characterized by disparate organizational structures, instruction sets, and annotation formats (e.g., varying between list-based and tag-based representations like \texttt{<box></box>}). Furthermore, spatial scales vary significantly, ranging from normalized $[0, 1]$ intervals to raw pixel dimensions. Such inconsistencies complicate the effective utilization of these resources. We think that GUI grounding is fundamentally a perception-centric task; therefore, optimization should prioritize spatial localization rather than being confounded by auxiliary capabilities like instruction following. To address this, we standardized all instructions into two primary categories (if the original task is GUI grounding task): bounding box prediction and center point localization (Prompt is given in \autoref{sec:prompt}). Additionally, we unified all spatial annotations to a $[0, 1]$ scale with three-decimal precision, formatted as consistent lists or tuples. Finally, the obtained homogeneous datasets are used to enhance the GUI grounding ability. \autoref{fig:fig3} presents examples of data samples reformatted from existing open-source datasets.

\paragraph{\textbf{Data Filtering}} The construction of existing open-source datasets typically follows two paradigms: leveraging high-performance models for screenshot annotation \citep{xie2025scalingcomputerusegroundinguser}, or utilizing attribute trees from web pages and applications to derive element coordinates \citep{wu2024atlas}. Since both approaches inherently introduce noise, we adopt the strategy of recent studies \citep{ye2025mobile, chen2025ui, zhou2025maiuitechnicalreportrealworld} by employing OmniParser-v2 \citep{lu2024omniparserpurevisionbased} to extract UI element coordinates, such as clickable text and icons. Our filtering pipeline handles two annotation formats: points and bounding boxes. For point-format annotations $(x, y)$, we first expand them into a square bounding box $B_{gt}$ with side length $l$. To validate these annotations, we calculate a coverage score $S$ relative to the set of $n$ elements $\mathcal{B}_{det} = \{B_{det, 1}, B_{det, 2}, \dots, B_{det, n}\}$ detected by OmniParser-v2. This metric is defined as:

\begin{equation}S = \frac{\sum_{i=1}^{n} \text{Area}(B_{gt} \cap B_{det, i})}{\text{Area}(B_{gt})}\end{equation}

where $B_{gt} \cap B_{det, i}$ denotes the intersection between the ground-truth annotation and the $i$-th detected UI element. We retain only samples where $S \geq \tau$, where $\tau$ represents a predefined reliability threshold. This refinement ensures our training data aligns strictly with perceivable visual structures, effectively pruning hallucinated or misaligned coordinates.

\begin{figure}[!htbp]
\centering
\includegraphics[width=\linewidth,scale=1.00]{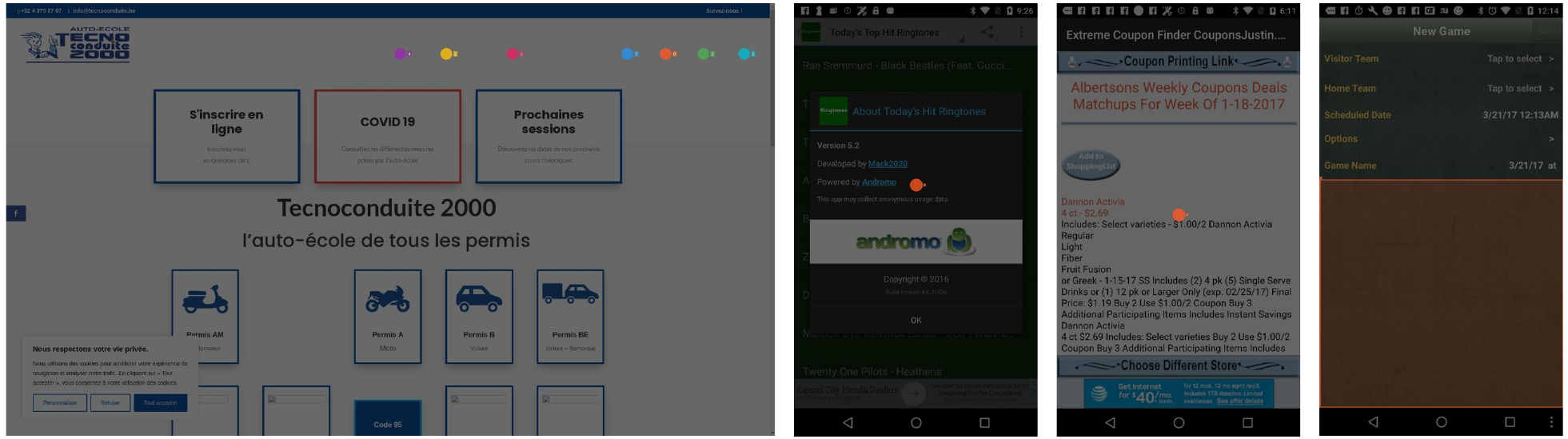}
\caption{Some samples filtered by our proposed filtering strategy. We place the original annotation onto the image.}
\label{fig:fig4}
\end{figure}

\begin{figure}[!htbp]
\centering
\includegraphics[width=\linewidth,scale=1.00]{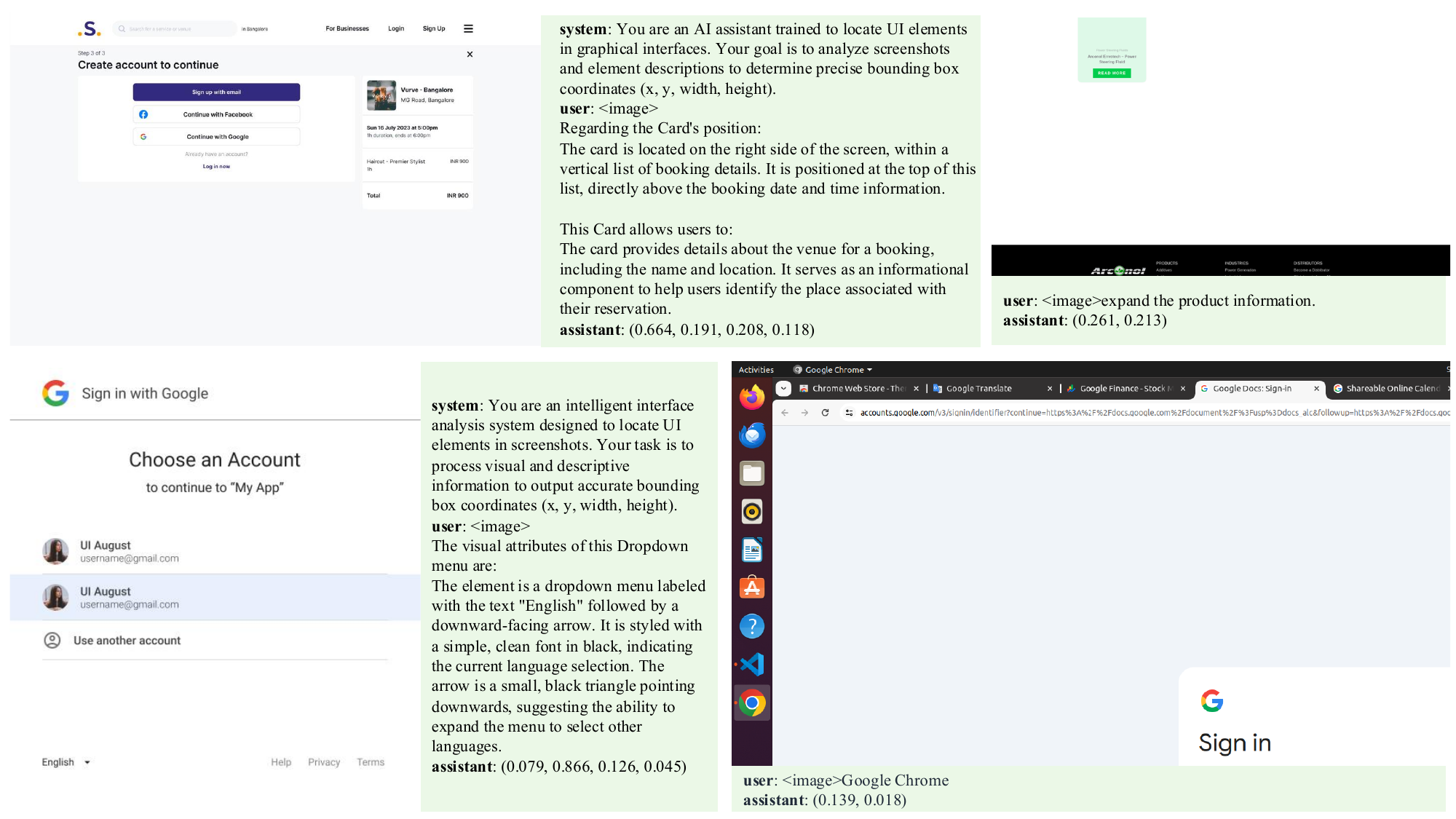}
\caption{Easy data samples from existing datasets filtered out by our complexity enhancement strategy.}
\label{fig:fig5}
\end{figure}

\paragraph{\textbf{Complexity Enhancement}} As models evolve, the original datasets often fail to sustain continuous performance gains. \autoref{fig:fig5} shows some easy samples from existing datasets. Consequently, it is essential to increase data complexity to drive more effective model learning. In this work, we propose a complexity filtering strategy based on layout entropy. Specifically, we define Layout Entropy, $E_\text{layout}$, to quantify the geometric complexity of a UI screen, containing a set of $N$ elements with location bounding boxes $\mathcal{B} = \{b_1, \dots, b_N\}$ within a normalized scale $[0,1]^2$. The total entropy is formulated as:
\begin{equation}
    E_\text{layout} = N^{w_N} \cdot \left( w_{1D} \bar{H}_{1D} + w_{2D} H_{2D} \right)
\end{equation}
where $w_N, w_{1D}, w_{2D}$ are weighting coefficients. $\bar{H}_{1D}$, ${H}_{2D}$ are defined as follows:

\begin{itemize}[leftmargin=*, labelsep=0.5em]
    \item \textbf{1D Projection Entropy ($\bar{H}_{1D}$):} This metric captures the alignment and distribution density of elements from multiple perspectives. Given $N$ bounding boxes with center points $\{(x_n, y_n)\}_{n=1}^N \in [0,1]^2$, we define $D$ uniform projection directions. Let the vertical axis be the reference direction ($\theta_1 = 0$), the $j$-th projection angle is defined as $\theta_j = \frac{(j-1)\pi}{D}$. For each direction $\theta_j$, we define a projection unit vector $\mathbf{u}_j = (\sin\theta_j, \cos\theta_j)$. The projection of the $n$-th center point onto this direction is calculated as:
    \begin{equation}
        z_{n,j} = x_n \sin\theta_j + y_n \cos\theta_j
    \end{equation}
    The projection direction is then partitioned into $B$ intervals (bins), denoted as $\{b_{1,j}, \dots, b_{B,j}\}$. The probability $p_{i,j}$ that a box falls into the $i$-th bin is:
    \begin{equation}
        p_{i,j} = \frac{1}{N} \sum_{n=1}^N \mathbb{I}(z_{n,j} \in b_{i,j})
    \end{equation}
    where $\mathbb{I}(\cdot)$ is the indicator function. The entropy for direction $\theta_j$ is $H_{1D}(\theta_j) = -\sum_{i=1}^B p_{i,j} \log p_{i,j}$, and the final 1D entropy is the average across all directions:
    \begin{equation}
        \bar{H}_{1D} = \frac{1}{D} \sum_{j=1}^D H_{1D}(\theta_j)
    \end{equation}

    \item \textbf{2D Grid Entropy ($H_{2D}$):} To evaluate the global spatial dispersion of elements, the screen is splitted into $M \times M$ grids. Similar to the probability estimation in the 1D projection, the probability $p_g$ for each cell $g$ is defined as the proportion of bounding box centers falling within its spatial boundaries:
    \begin{equation}
        p_g = \frac{1}{N} \sum_{n=1}^N \mathbb{I}((x_n, y_n) \in \text{cell}_g)
    \end{equation}
    The 2D spatial entropy is then formulated as:
    \begin{equation}
        H_{2D} = -\sum_{g=1}^{G} p_g \log p_g
    \end{equation}
    A lower $H_{2D}$ indicates that elements are concentrated in a few grid cells, while a higher value signifies a more uniform and complex distribution across the interface.
\end{itemize}

After calculating the Layout Entropy $E_\text{layout}$ for all data, we utilize this metric to categorize the dataset into three difficulty levels (easy, medium, and hard).

For \textbf{image resolution}, we differentiate samples based on their total pixels $S = W \times H$. High resolution samples are prioritized as they typically contain higher information density and fine-grained visual details.

Regarding \textbf{data synthesis}, we introduce two strategies designed to increase task complexity. 1) \textbf{GUI-CodeGen}: We leverage the sophisticated layouts of professional software—such as VS Code—which are characterized by a high density of functional components. Utilizing Large Language Models (e.g., Claude), we synthesize the underlying frontend HTML for these interfaces, extract operable elements like icons and input fields, and render them into high-resolution images (e.g. $1920 \times 2560$). 2) \textbf{GUI-Overlay}: To simulate real-world scenarios, we overlay multiple application windows onto diverse desktop backgrounds. This method introduces significant visual distractors and occlusions, challenging the model’s ability to accurately identify and interact with target elements amidst background interference.

\subsection{Training Strategies}
\label{sec:training_strategies}
\paragraph{\textbf{Adapting Vision Encoder to GUI Scenario}} Historically, the WePOINTS series—including POINTS-Reader~\citep{liu-etal-2025-points} and POINTS-1.5~\citep{liu2024points1}—maintained a frozen vision encoder, which yielded strong performance across various tasks. However, we observed that for GUI grounding, existing encoders such as Qwen2-VL-NaViT~\citep{wang2024qwen2} are no longer sufficient. This limitation likely stems from the relative scarcity of GUI-specific data during the encoder's initial pre-training. To address this, we unfreeze the vision encoder throughout the training process, which resulted in substantial improvements in model performance.

\paragraph{\textbf{Maintaining Resolution Consistency}} To optimize memory usage and training efficiency, we initially capped the training image resolution at $2000 \times 2000$ pixels. However, no such constraints were applied during inference. We observed that this resolution discrepancy led to suboptimal performance on benchmarks featuring high-resolution images, such as ScreenSpot-Pro~\citep{li2025screenspotpro}. Drawing on prior research regarding the impact of resolution consistency~\citep{liu2023pixmim, liu2023improving, he2022masked, dosovitskiy2020image}, we investigated two adjustments to our pipeline: (1) capping inference resolution at $2000 \times 2000$ to align with training, and (2) increasing the training resolution limit to $2500 \times 2500$. Our experiments demonstrate that both strategies significantly improve model performance by mitigating the train-test inconsistency.

\subsection{Reinforcement Learning with Verifiable Rewards}
\label{sec:rlvr}
\paragraph{\textbf{Training Data Construction}} Following the pipeline described in \autoref{sec:data_engineer}, we construct training samples by overlaying multiple application windows onto desktop backgrounds. To facilitate more effective learning and mitigate zero-gradient issues, we use the initialized model to perform eight rollouts per task. We retain only those samples with a pass rate between 0\% and 75\%, similar to recent work~\citep{yu2025dapo}.

\paragraph{\textbf{Training Algorithm}} We employ Group Relative Policy Optimization (GRPO) \citep{shao2024deepseekmath}, optimizing the objective function:

\begin{equation}
\mathcal{J}_{\text{GRPO}}(\theta) = \mathbb{E}_{q \sim \mathcal{D}, \{o_i\}_{i=1}^G \sim \pi_{\theta_{\text{old}}}} \left[ \frac{1}{\sum_{c=1}^G |o_c|} \sum_{i=1}^{G} \sum_{t=1}^{|o_i|} \min \left( r_{i,t}(\theta) \hat{A}_{i,t}, \text{clip} \left( r_{i,t}(\theta), 1 - \varepsilon, 1 + \varepsilon \right) \hat{A}_{i,t} \right) \right] 
\end{equation}

where $r_{i,t}(\theta) = \frac{\pi_{\theta}(o_{i,t}|q, o_{i,<t})}{\pi_{\theta_{\text{old}}}(o_{i,t}|q, o_{i,<t})}$ is the importance sampling ratio, and $\hat{A}_{i,t} = \frac{R_i - \text{mean}(\{R_j\}_{j=1}^G)}{\text{std}(\{R_j\}_{j=1}^G)}$ is the group-normalized advantage. We found that a group size of $G=8$ strikes an optimal balance between training effectiveness and efficiency. A key advantage of the GUI grounding task is that it permits a precise, verifiable reward function. Specifically, the reward $R_i$ for a predicted coordinate $(x_n, y_n)$ is binary, determined by whether it falls within the annotated bounding box $b^{\text{ann}} = (x_{\min}, y_{\min}, x_{\max}, y_{\max})$:

\begin{equation}
    R_i = \begin{cases} 1, & \text{if } x_{\min} \le x_n \le x_{\max} \text{ and } y_{\min} \le y_n \le y_{\max}, \\ 0, & \text{otherwise.} \end{cases}
\end{equation}
\section{Experiments Settings}
\paragraph{\textbf{Training}} Building upon the architecture of POINTS-1.5~\citep{liu2024points1}, we replace the Qwen2.5-7B-Instruct backbone~\citep{Yang2024Qwen25TR} with Qwen3-8B~\citep{qwen3technicalreport}. Prior to specialized optimization for GUI tasks, the model underwent extensive pre-training and mid-training~\citep{Bai2025Qwen3VLTR,Guo2025Seed15VLTR,Hong2025GLM45VAG} on large-scale datasets. The optimization of the GUI Grounding ability is diveded into two stages: supervised learning stage and reinforcement learning stage. During the supervised learning stage, we jointly fine-tune all components, including the vision encoder, projector, and large language model (LLM). We set the learning rate to $1 \times 10^{-4}$ for the vision encoder and $5 \times 10^{-5}$ for the projector and LLM. For the reinforcement learning stage, we employ 8 rollouts per sample with a global batch size of 64 and a learning rate of $1 \times 10^{-5}$.

\begin{figure}[!htbp]
\centering
\includegraphics[width=\linewidth,scale=1.00]{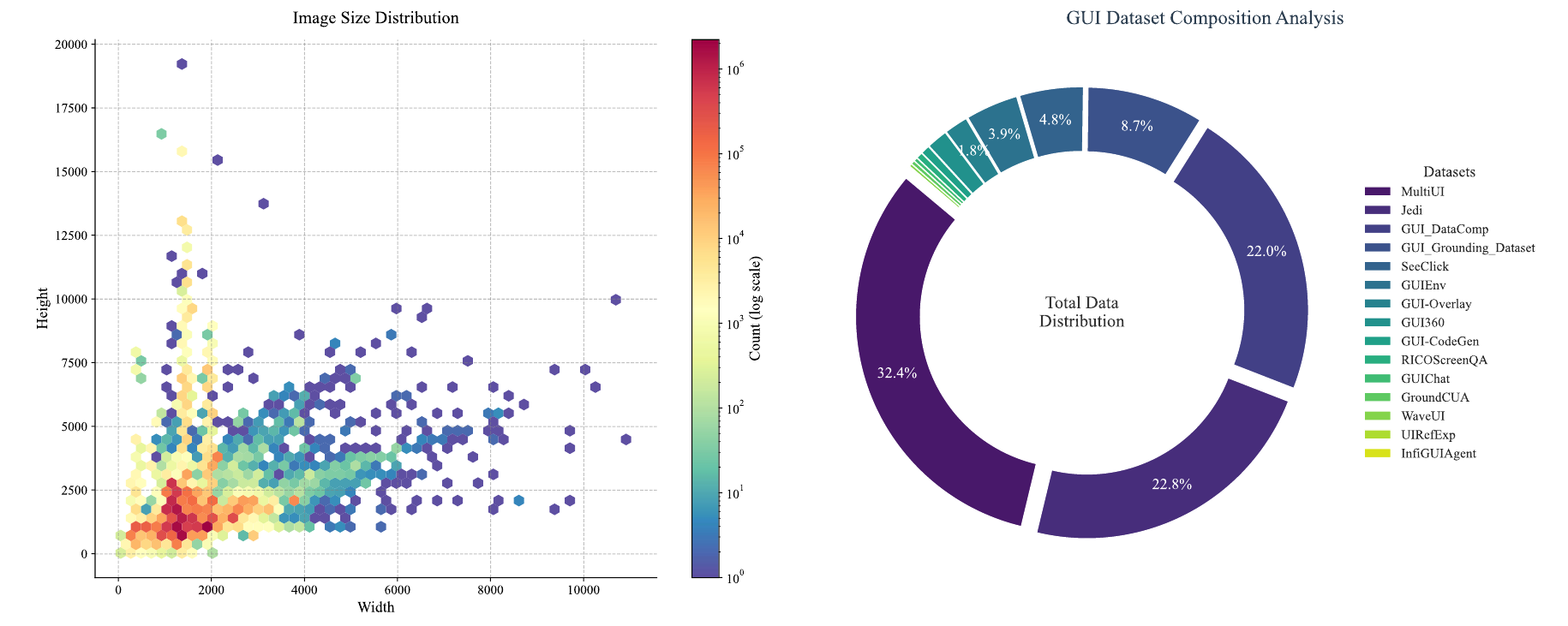}
\caption{\textbf{Data distribution of all these GUI grounding datasets.} Left: image shape distribution. Right: composition of these GUI Grounding datasets.}
\label{fig:fig6}
\end{figure}
\begin{table*}[!ht]
    \centering
    \small
    \caption{Performance comparison on \textbf{ScreenSpot-V2}. The best results are highlighted in \textbf{bold}, and the second-best results are \underline{underlined}.}
    \label{tab:ssv2}

    \setlength{\tabcolsep}{3pt} 
    
    \begin{tabularx}{\textwidth}{
        l
        *{6}{>{\centering\arraybackslash}X}
        c
        }
        \toprule
        \multirow{2}{*}{\textbf{Model}} 
        & \multicolumn{2}{c}{\textbf{Mobile}} &
          \multicolumn{2}{c}{\textbf{Desktop}} &
          \multicolumn{2}{c}{\textbf{Web}} &
          \multirow{2}{*}{\textbf{Avg.}}  \\
        \cmidrule(lr){2-3} \cmidrule(lr){4-5} \cmidrule(lr){6-7}
        
        & Text & Icon. & Text & Icon. & Text & Icon. &  \\
        \midrule
        \rowcolor{gray!15}
        \multicolumn{8}{l}{\textit{Open-Source Models}} \\
        Qwen3-VL-2B~\citep{Bai2025Qwen3VLTR} &95.5 &82.0 &95.4 &73.6 &89.7 &76.4 &\cellcolor[HTML]{d8f3dc}86.7 \\
        Phi-ground~\citep{phi_ground}
            & 90.2 & 76.4 & 93.6 & 75.9 & \underline{96.5} & 62.0
            & \cellcolor[HTML]{d8f3dc}83.8
            \\
        OS-Atlas-7B~\citep{wu2024atlas}
            & 95.2 & 75.8 & 90.7 & 63.6 & 90.6 & 77.3
            & \cellcolor[HTML]{d8f3dc}85.1
           \\
        UGround-v1-7B~\citep{uground}
            & 83.6 & 90.5 & 85.8 & 86.3 & 95.5 & 83.2
            & \cellcolor[HTML]{d8f3dc}87.7
            \\
        UI-Tars-1.5-7B~\citep{ui-tars-15-seed}
            & 92.2 & 81.5 & 91.0 & 84.2 & 95.5 & 84.5
            & \cellcolor[HTML]{d8f3dc}89.0
            \\

        SE-GUI-7B~\citep{SE_GUI}
            & \underline{99.3} & 89.1 &
              96.4 & 78.6 &
              92.7 & 81.3 &
              \cellcolor[HTML]{d8f3dc}90.8 
              \\
        UI-TARS-7B~\citep{qin2025ui}
            & 96.9 & 89.1 & 95.4 & 85.0 & 93.6 & 85.2
            & \cellcolor[HTML]{d8f3dc}91.6
             \\
        Qwen3-VL-8B~\citep{Bai2025Qwen3VLTR} &97.9 &84.8 &95.9 &87.9 &95.7 &83.7 &\cellcolor[HTML]{d8f3dc}91.7\\
        GUI-Actor-7B~\citep{wu2025gui}
            & 97.6 & 88.2 & 96.9 & 85.7 & 93.2 & 86.7
            & \cellcolor[HTML]{d8f3dc}92.1
             \\
        OpenCUA-7B~\citep{opencua}
            & - & - & - & - & - & -
            & \cellcolor[HTML]{d8f3dc}92.3
            \\
        GTA1-7B~\citep{yang2025gta1}
            & 99.0 & 88.6 & 94.9 & 89.3 & 92.3 & 86.7
            & \cellcolor[HTML]{d8f3dc}92.4
             \\
        GUI-Owl-7B~\citep{ye2025mobile}
            & 99.0 & \underline{92.4} & 96.9 & 85.0 & 93.6 & 85.2 & 
            \cellcolor[HTML]{d8f3dc}92.8 \\
        GUI-G$^2$-7B~\citep{GUI_G2}
            & 98.3 & 91.9 &
              95.4 & 89.3 &
              94.0 & 87.7 &
              \cellcolor[HTML]{d8f3dc}93.3 
              \\
        InfiGUI-G1-7B~\citep{infiguig1}
            & 99.0 & 91.9 &
              94.3 & 82.1 &
              \textbf{97.9} & 89.2 &
              \cellcolor[HTML]{d8f3dc}93.5 
              \\
        UI-Venus-7B~\citep{ui_venus}
            & 99.0 & 90.0 &
              96.9 & 90.7 &
              96.2 & 88.7 &
              \cellcolor[HTML]{d8f3dc}94.1
               \\
        
        MAI-UI-2B~\citep{zhou2025maiuitechnicalreportrealworld}& \underline{99.3} &87.2 &97.4 &88.6 &94.0 &84.7 &\cellcolor[HTML]{d8f3dc}92.5\\
        
        MAI-UI-8B~\citep{zhou2025maiuitechnicalreportrealworld}&\underline{99.3} &89.1 &\underline{99.0} &\underline{92.1} &\textbf{97.9} &\underline{91.1} &\cellcolor[HTML]{d8f3dc}95.2\\
        \midrule
        Qwen3-VL-32B~\citep{Bai2025Qwen3VLTR} & 96.2 &90.0 &97.4 &85.0 &95.7 &89.7 &\cellcolor[HTML]{d8f3dc}93.0\\
        GUI-Owl-32B~\citep{ye2025mobile}
            & 98.6 & 90.0 & 97.9 & 87.8 & 94.4 & 86.7 & 
            \cellcolor[HTML]{d8f3dc}93.2 \\
        OpenCUA-32B~\citep{opencua}
            & - & - & - & - & - & -
            & \cellcolor[HTML]{d8f3dc}93.4
            \\

        GTA1-32B~\citep{yang2025gta1}
            & \textbf{99.7} & 90.5 & \underline{99.0} & \textbf{94.3} & 95.7 & 90.1
            & \cellcolor[HTML]{d8f3dc}95.2
            \\
        UI-Venus-72B~\citep{ui_venus}
            & \textbf{99.7} & \textbf{93.8} & 95.9 & 90.0 & 96.2 & \textbf{92.6} & 
            \cellcolor[HTML]{d8f3dc}\underline{95.3} \\
        \midrule
        \rowcolor{gray!15}
        \multicolumn{8}{l}{\textit{Ours}} \\

        \textbf{POINTS-GUI-G-8B}&99.0 &91.0 &\textbf{100.0} &\textbf{94.3} &95.3 &\textbf{92.6} &\cellcolor[HTML]{d8f3dc}\textbf{95.7}\\
        \bottomrule
    \end{tabularx}
\end{table*}
\begin{table}[!ht]
    \caption{Performance comparison of state-of-the-art models on the \textbf{OSWorld-G}. The best results are highlighted in \textbf{bold}, and the second-best results are \underline{underlined}.}
    \centering
    \setlength{\tabcolsep}{4pt}
    \small
    \begin{tabular}{lcccccc}
        \toprule
        \textbf{Agent Model} & \makecell[c]{\textbf{Text}\\\textbf{Matching}} & \makecell[c]{\textbf{Element}\\\textbf{Recognition}} & \makecell[c]{\textbf{Layout}\\\textbf{Understanding}} & \makecell[c]{\textbf{Fine-grained}\\\textbf{Manipulation}}& \textbf{Refusal} & \cellcolor{white}\textbf{Avg} \\
        \midrule
        \rowcolor{gray!15}
        \multicolumn{7}{l}{\textit{Proprietary Models}} \\
        Operator  \citep{OpenAICUA} & 51.3 & 42.4 & 46.6 & 31.5 & 0.0 & \cellcolor[HTML]{d8f3dc}40.6 \\
        Seed1.5-VL \citep{Guo2025Seed15VLTR} & \underline{73.9} & 66.7 & 69.6 & 47.0 & \textbf{18.5} &\cellcolor[HTML]{d8f3dc}62.9 \\
        \midrule
        \rowcolor{gray!15}
        \multicolumn{7}{l}{\textit{Open-Source Models}} \\
        Jedi-3B \citep{xie2025scalingcomputerusegroundinguser} & 67.4 & 53.0 & 53.8 & 44.3 & \underline{7.4} & \cellcolor[HTML]{d8f3dc}50.9 \\
        Qwen3-VL-2B~\citep{Bai2025Qwen3VLTR}& 61.7 &45.8 &54.2 &39.6 &- &\cellcolor[HTML]{d8f3dc}45.9\\
        OS-Atlas-7B \citep{wu2024atlas}& 44.1 & 29.4 & 35.2 & 16.8 & \underline{7.4} & \cellcolor[HTML]{d8f3dc}27.7 \\
        UGround-7B \citep{uground} & 51.3 & 40.3 & 43.5 & 24.8 & 0.0 & \cellcolor[HTML]{d8f3dc}36.4 \\
        Aguvis-7B \citep{Xu2024AguvisUP} & 55.9 & 41.2 & 43.9 & 28.2 & 0.0 & \cellcolor[HTML]{d8f3dc}38.7 \\
        UI-TARS-7B \citep{qin2025ui} & 60.2 & 51.8 & 54.9 & 35.6 & 0.0 & \cellcolor[HTML]{d8f3dc}47.5 \\
        UI-TARS-1.5-7B \citep{ui-tars-15-seed} & 36.8 & 62.7 & 62.2 & 50.8 & 0.0 & \cellcolor[HTML]{d8f3dc}52.8\\
        Jedi-7B \citep{xie2025scalingcomputerusegroundinguser} & 65.9 & 55.5 & 57.7 & 46.9 & \underline{7.4} & \cellcolor[HTML]{d8f3dc}54.1 \\
        Qwen3-VL-8B~\citep{Bai2025Qwen3VLTR}&69.0 &55.5 &59.7 &47.7 &- &\cellcolor[HTML]{d8f3dc}54.8\\
        GTA1-7B~\citep{yang2025gta1} & 42.1 & 65.7 & 62.7 & \underline{56.1} & 0.0 & \cellcolor[HTML]{d8f3dc}55.1 \\
        GUI-Owl-7B~\citep{ye2025mobile} & 64.8 & 63.6 & 61.3 & 41.0 & - & \cellcolor[HTML]{d8f3dc}55.9 \\
        UI-Venus-7B~\citep{ui_venus} & \textbf{74.6} & 60.5 & 61.5 & 45.5 & - & \cellcolor[HTML]{d8f3dc}58.8 \\

        MAI-UI-2B~\citep{zhou2025maiuitechnicalreportrealworld} & 62.8 &56.7 &59.3 &40.3 &- & \cellcolor[HTML]{d8f3dc}52.0\\
        
        MAI-UI-8B~\citep{zhou2025maiuitechnicalreportrealworld} & 72.0 &63.3 &66.0 &51.0 &- & \cellcolor[HTML]{d8f3dc}60.1\\
        \midrule
        OpenCUA-32B \citep{opencua} & - & - & - & - & - & \cellcolor[HTML]{d8f3dc}59.6 \\
        GUI-Owl-32B~\citep{ye2025mobile} & 67.0 & 64.5 & 67.2 & 45.6 & - & \cellcolor[HTML]{d8f3dc}58.0 \\
        Qwen3-VL-32B~\citep{Bai2025Qwen3VLTR}&72.8 &63.3 &66.4 &51.7 &- &\cellcolor[HTML]{d8f3dc}60.6\\

        GTA1-32B~\citep{yang2025gta1} & 63.2 & \textbf{78.4} & \textbf{73.3} & \textbf{65.2} &0.0&\cellcolor[HTML]{d8f3dc}\underline{65.2}  \\
        \rowcolor{gray!15}
        \multicolumn{7}{l}{\textit{Ours}} \\
        
        \textbf{POINTS-GUI-G-8B} & \underline{73.9} &\underline{73.6} &\underline{70.8} &55.7 &- & \cellcolor[HTML]{d8f3dc}\textbf{66.0}\\
        \bottomrule
    \end{tabular}
    \label{tab:osworld_g}
    \vspace{-0.8em}
\end{table}

\paragraph{\textbf{Datasets}} 
In addition to the dataset synthesized via the strategies detailed in \autoref{sec:data_engineer}, we curate a text-centric subset from DataComp~\citep{gadre2023datacomp} by filtering for images containing textual elements. We employ PaddleOCR~\citep{cui2025paddleocr30technicalreport} to extract text bounding boxes, utilizing the text as the query and the coordinates as the ground-truth answer (\textbf{GUI-DataComp}). To further enhance diversity, we incorporate 13 additional open-source datasets~\citep{liu2024harnessingwebpageuistextrich, xie2025scalingcomputerusegroundinguser, hellokkme_grounding_dataset, cheng2024seeclick, chen2025guicourse, mu2025gui, hsiao2024screenqa, feizi2025groundingcomputeruseagents, waveui, uirefexp, liu2025infiguiagent}. The geometric and numerical distributions of these GUI datasets are visualized in \autoref{fig:fig6}. Beyond specialized GUI data, our training pipeline integrates general-purpose corpora, such as Bee~\citep{zhang2025beehighqualitycorpusfullstack}. Performance is evaluated across five commonly-used benchmarks: ScreenSpot-v2~\citep{wu2024atlas}, ScreenSpot-Pro~\citep{li2025screenspotpro}, OSWorld-G~\citep{xie2025scalingcomputerusegroundinguser}, MMBench-GUI-L2~\citep{Wang2025MMBenchGUIHM}, and UI-Vision~\citep{Nayak2025UIVisionAD}. Together, these benchmarks provide a comprehensive assessment across mobile, web, and desktop environments.
\section{Experiment Results}
\begin{figure}[!htbp]
\centering
\includegraphics[width=\linewidth,scale=0.95]{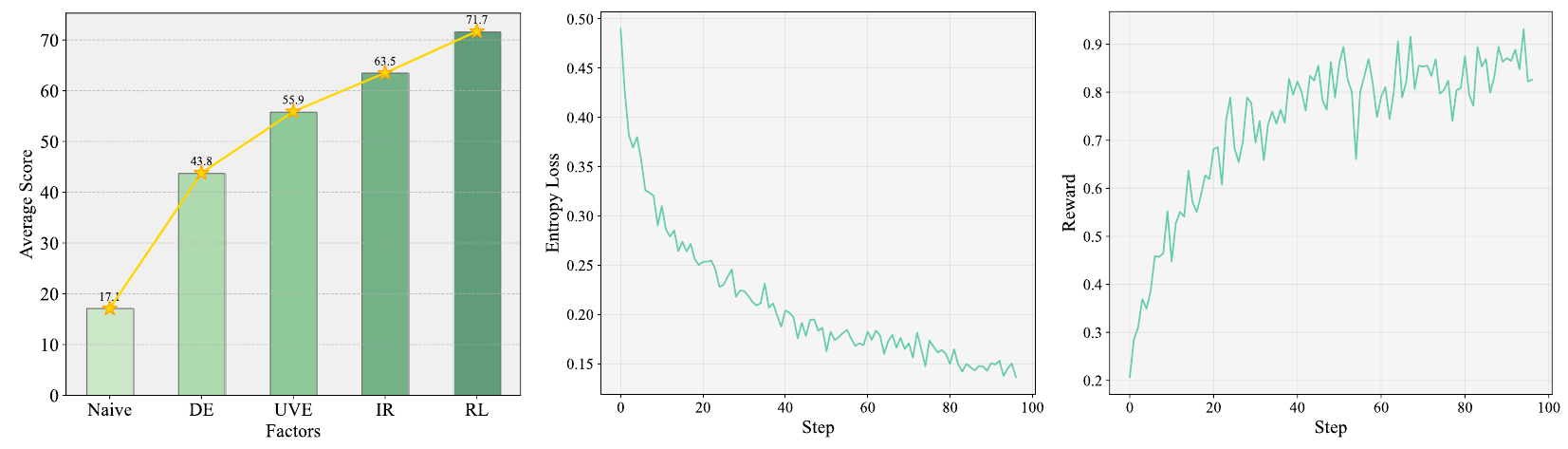}
\caption{\textbf{Analysis of influential factors and the RL dynamics.} \textbf{Left}: Comparison of key factors affecting performance (average score across the five evaluation benchmarks we use). \textbf{Middle \& Right}: The Reinforcement Learning (RL) training procedure. \textbf{Naive}: baseline without GUI grounding optimization; \textbf{DE}: Data Engineering (see \autoref{sec:data_engineer}); \textbf{UVE}: Unfrozen Vision Encoder; \textbf{IR}: Image Resolution consistency; \textbf{RL}: Reinforcement Learning.}
\label{fig:fig7}
\end{figure}
\paragraph{\textbf{The Most Influential Factors}} The development of large-scale models necessitate the integration of diverse technologies, including data curation and training strategies. This evolutionary process is often characterized by distinct plateau periods, where specific pivotal factors determine the transition to higher performance levels. \autoref{fig:fig4}(left) illustrates the key milestones that proved decisive in the iteration of POINTS-GUI-G. Notably, the data presented reflects the aggregate average performance of all models within each plateau period, rather than the performance of a single individual model.

\paragraph{\textbf{The Training Procedure of Reinforcement Learning}} Due to the high variance and instability characteristic of RL training, tracking intermediate performance via Reward and Entropy Loss is crucial. Reward serves as a proxy for policy improvement, whereas Entropy Loss characterizes the model's exploration-exploitation trade-off and the sharpening of its output distribution. Our experimental results (\autoref{fig:fig4}(right)) show a consistent rise in reward followed by a stable plateau. Simultaneously, the fluctuating decrease in entropy loss demonstrates that the model continues to explore while progressively increasing the likelihood of generating optimal tokens (\autoref{fig:fig4}(middle)).

\begin{table}[ht!]
\centering
\footnotesize
\small
\setlength{\tabcolsep}{16pt}
\caption{Performance comparison on \textbf{UI-Vision} grounding dataset. The best results are highlighted in \textbf{bold}, and the second-best results are \underline{underlined}.}

\begin{tabular}{lcccc}
\toprule
\textbf{Models} & \textbf{Basic} & \textbf{Functional} & \textbf{Spatial}& \textbf{Avg}\\ 
\midrule
\rowcolor{gray!15}
\multicolumn{5}{l}{\textit{Proprietary Models}} \\
GPT-4o~\citep{gpt4o}& 1.6 & 1.5 &1.0&\cellcolor[HTML]{d8f3dc}1.4\\ 
Claude-3.7-Sonnet~\citep{claude37}& 9.5 & 7.7&7.6&\cellcolor[HTML]{d8f3dc}8.3 \\
\midrule
\rowcolor{gray!15}
\multicolumn{5}{l}{\textit{Open-source Models}} \\
Qwen3-VL-2B~\citep{Bai2025Qwen3VLTR} &0.0 &19.2 &0.1 &\cellcolor[HTML]{d8f3dc}6.2 \\
InfiGUI-G1-3B~\citep{infiguig1} & 31.2 & 28.0 & 8.2 & \cellcolor[HTML]{d8f3dc}22.0\\
Qwen2.5-VL-7B~\citep{Qwen2.5-VL}&1.2& 0.8&0.5&\cellcolor[HTML]{d8f3dc}0.9\\ 
SeeClick~\citep{cheng2024seeclick}&9.4 & 4.7&2.1&\cellcolor[HTML]{d8f3dc}5.4 \\ 
UGround-V1-7B~\citep{uground}&15.4& 17.1& 6.3 &\cellcolor[HTML]{d8f3dc}12.9\\ 
OS-Atlas-7B~\citep{wu2024atlas}&12.2 &11.2 &3.7 &\cellcolor[HTML]{d8f3dc}9.0\\
Qwen3-VL-8B~\citep{Bai2025Qwen3VLTR} &25.0 &27.9 &1.2 &\cellcolor[HTML]{d8f3dc}17.5 \\
UI-TARS-7B~\citep{qin2025ui}& 20.1 &24.3 &8.4 &\cellcolor[HTML]{d8f3dc}17.6\\

UI-TARS-1.5-7B~\citep{ui-tars-15-seed}& 28.8 &27.5& 10.7 &\cellcolor[HTML]{d8f3dc}22.3 \\
InfiGUI-G1-7B~\citep{infiguig1} & 36.2 & 31.9 & 11.5 & \cellcolor[HTML]{d8f3dc}26.1 \\
UI-Venus-7B~\citep{ui_venus} &36.1&32.8 &11.9&\cellcolor[HTML]{d8f3dc}26.5\\ 
Phi-Ground~\citep{phi_ground}&36.8 &37.1 & 7.6 &\cellcolor[HTML]{d8f3dc}27.2\\
MAI-UI-2B~\citep{zhou2025maiuitechnicalreportrealworld} & 41.0 &41.2 &10.4 &\cellcolor[HTML]{d8f3dc}30.3\\
MAI-UI-8B~\citep{zhou2025maiuitechnicalreportrealworld} & \underline{51.7} &\underline{49.6} &22.5 & \cellcolor[HTML]{d8f3dc}\underline{40.7}\\
\midrule
Qwen3-VL-32B~\citep{Bai2025Qwen3VLTR} & 32.8 &34.2 &14.7 &\cellcolor[HTML]{d8f3dc}26.9\\
UI-TARS-72B~\citep{qin2025ui}& 31.4 &30.5 & 14.7 & \cellcolor[HTML]{d8f3dc}25.5\\
UI-Venus-72B~\citep{ui_venus}& 45.6 & 42.3 & \underline{23.7}&\cellcolor[HTML]{d8f3dc}36.8\\

\midrule
\rowcolor{gray!15}
\multicolumn{5}{l}{\textit{Ours}} \\
\textbf{POINTS-GUI-G-8B} & \textbf{63.2} &\textbf{55.6} &\textbf{30.9} & \cellcolor[HTML]{d8f3dc}\textbf{49.9}\\
\bottomrule
\end{tabular}
\label{tab:ui-vison}
\vspace{-0.8em}
\end{table}

\begin{table*}[ht!]
    \centering
    \small
    \setlength{\tabcolsep}{1.6pt}
    \caption{Performance comparison on the \textbf{MMBench-GUI L2} benchmark. The best results are highlighted in \textbf{bold}, and the second-best results are \underline{underlined}.}
    \label{tab:mmbench-gui}

    \begin{tabularx}{\textwidth}{
        l
        *{12}{>{\centering\arraybackslash}X}
        c}
        \toprule
        \multirow{2}{*}{\textbf{Model}} &
        \multicolumn{2}{c}{\textbf{Windows}} &
        \multicolumn{2}{c}{\textbf{MacOS}} &
        \multicolumn{2}{c}{\textbf{Linux}} &
        \multicolumn{2}{c}{\textbf{iOS}} &
        \multicolumn{2}{c}{\textbf{Android}} &
        \multicolumn{2}{c}{\textbf{Web}} &
        \multirow{2}{*}{\textbf{Avg.}} \\
        \cmidrule(lr){2-3}
        \cmidrule(lr){4-5}
        \cmidrule(lr){6-7}
        \cmidrule(lr){8-9}
        \cmidrule(lr){10-11}
        \cmidrule(lr){12-13}
        & Bas. & Adv. & Bas. & Adv. & Bas. & Adv. & Bas. & Adv. & Bas. & Adv. & Bas. & Adv. & \\
        \midrule
        \rowcolor{gray!15}
        \multicolumn{14}{l}{\textit{Proprietary Models}} \\
        GPT-4o~\citep{gpt4o}
            & 1.5 & 1.1 & 8.7 & 4.3 & 1.1 & 1.0 & 5.1 & 3.3 & 2.5 & 1.4 & 3.2 & 2.9
            & \cellcolor[HTML]{d8f3dc}2.9 \\
        Claude-3.7~\citep{claude37}
            & 1.5 & 0.7 & 12.5 & 7.5 & 1.1 & 0.0 & 13.7 & 10.6 & 1.4 & 1.4 & 3.2 & 2.3
            & \cellcolor[HTML]{d8f3dc}4.7 \\
        Qwen-Max-VL~\citep{qwen2}
            & 43.9 & 36.8 & 58.8 & 56.1 & 53.9 & 30.1 & 77.4 & 59.1 & 79.5 & 70.1 & 74.8 & 58.8
            & \cellcolor[HTML]{d8f3dc}58.0 \\
        \midrule
        \rowcolor{gray!15}
        \multicolumn{14}{l}{\textit{Open-Source Models}} \\
        Qwen3-VL-2B~\citep{Bai2025Qwen3VLTR}
            & 81.9 &0.0 &80.3 &46.2 &67.5 &0.0 &90.8 &0.0 &91.0 &0.0 &88.1 &0.0 &\cellcolor[HTML]{d8f3dc}46.5\\
        OS-Atlas-7B~\citep{wu2024atlas}
            & 36.9 & 18.8 & 44.4 & 21.7 & 31.4 & 13.3 & 74.8 & 48.8 & 69.6 & 46.8 & 61.3 & 35.4
            & \cellcolor[HTML]{d8f3dc}41.4 \\
        Aguvis-7B~\citep{Xu2024AguvisUP}
            & 37.3 & 21.7 & 48.1 & 33.3 & 33.5 & 25.0 & 67.5 & 65.2 & 61.0 & 51.0 & 61.6 & 45.5
            & \cellcolor[HTML]{d8f3dc}45.7 \\
        UI-TARS-1.5-7B~\citep{ui-tars-15-seed}
            & 68.3 & 39.0 & 69.0 & 44.5 & 64.4 & 37.8 & 88.5 & 69.4 & 90.5 & 69.3 & 81.0 & 56.5
            & \cellcolor[HTML]{d8f3dc}64.3 \\
        UGround-V1-7B~\citep{Gou2024NavigatingTD}
            & 66.8 & 39.0 & 71.3 & 48.6 & 56.5 & 31.1 & 92.7 & 70.9 & 93.5 & 71.0 & 88.7 & 64.6
            & \cellcolor[HTML]{d8f3dc}65.7 \\
        GUI-Actor-7B~\citep{wu2025gui}
            & 80.8 & 55.1 & 81.4 & 60.4 & 64.9 & 41.8 & 94.3 & 82.7 & 93.5 & 79.7 & 89.7 & 72.1
            & \cellcolor[HTML]{d8f3dc}76.5 \\
        SE-GUI-7B~\citep{SE_GUI}
            & 77.5 & 57.7 & 77.1 & 60.7 & 68.6 & 44.9 & 95.5 & 80.0 & 95.5 & 83.7 & 89.7 & 68.8
            & \cellcolor[HTML]{d8f3dc}76.6 \\
        Qwen3-VL-8B~\citep{Bai2025Qwen3VLTR} 
            & 88.6 &61.8 &85.5 &69.1 &74.9 &53.1 & 95.2 &82.4 &95.5 &84.5 & \textbf{96.8} &72.1 &\cellcolor[HTML]{d8f3dc}81.3\\
        GTA1-7B~\citep{yang2025gta1}
            & 76.8 & 57.4 & 80.3 & 63.9 & 68.6 & 53.6 & 93.9 & 83.3 & 96.3 & 84.5 & 90.3 & 74.7
            & \cellcolor[HTML]{d8f3dc}78.5 \\
        GUI-G$^2$-7B~\citep{GUI_G2}
            & 79.7 & 55.1 & 79.7 & 64.7 & 69.6 & 50.0 & 95.2 & 82.7 & 96.6 & 85.4 & 91.9 & 75.6
            & \cellcolor[HTML]{d8f3dc}78.8 \\
        GUI-Owl-7B ~\citep{ye2025mobile} 
            & 86.4 & 61.8 & 81.7 & 64.5 & 74.4 & 61.7
            & 94.9 & 83.0 & 95.8 & 83.7 & 93.2 & 72.7
            & \cellcolor[HTML]{d8f3dc}80.5 \\
        InfiGUI-G1-7B~\citep{infiguig1}
            & 82.7 & 61.8 & 83.8 & 63.9 & 72.3 & 52.0 & 94.9 & \underline{89.4} & 95.2 & 85.6 & 93.5 & 76.3
            & \cellcolor[HTML]{d8f3dc}80.8 \\
        
        MAI-UI-2B~\citep{zhou2025maiuitechnicalreportrealworld} &84.9 &64.0 &89.3 &72.5 &75.4 &60.2 &95.2 &85.2 &96.3 &84.2 &92.9 &76.0 & \cellcolor[HTML]{d8f3dc}82.6 \\
        
        MAI-UI-8B~\citep{zhou2025maiuitechnicalreportrealworld}& 92.3 & \textbf{74.3} & \underline{90.7} & \textbf{86.4} & \underline{81.2} & \underline{67.3} & \textbf{97.1} & \textbf{90.0} & \textbf{97.5} & \textbf{92.7} & \underline{95.8} & \textbf{86.0} & \cellcolor[HTML]{d8f3dc}\textbf{88.8}\\

        \midrule


        GUI-Owl-32B ~\citep{ye2025mobile} 
            & 85.6 & 65.1 & 84.9& 67.1 & 77.0 & 63.3
            & 95.2 & 85.5 & 96.1 & 87.0 & 95.5 & 80.8
            & \cellcolor[HTML]{d8f3dc}83.0 \\
        GTA1-32B~\citep{yang2025gta1}
            & 82.3 & 66.9 & 89.0 & 74.0 & 73.3 & 52.0
            & \underline{96.2} & 88.2 & 95.8 & 88.5 & 95.2 & 79.9
            & \cellcolor[HTML]{d8f3dc}83.4 \\
        Qwen3-VL-32B~\citep{Bai2025Qwen3VLTR} 
            & \underline{93.4} & \underline{71.3} & \textbf{92.8} & \underline{74.3} &78.0 &56.1 &95.5 &88.8 & \underline{97.2} &88.5 &92.6 &78.6 &\cellcolor[HTML]{d8f3dc}85.3\\

        UI-TARS-DPO-72B~\citep{qin2025ui}
            & 78.6 & 51.8 & 80.3 & 62.7 & 68.6 & 51.5 & 90.8 & 81.2 & 93.0 & 80.0 & 88.1 & 68.5
            & \cellcolor[HTML]{d8f3dc}74.3 \\
        InternVL3-78B~\citep{zhu2025internvl3}
            & 70.1 & 42.6 & 75.7 & 52.3 & 59.2 & 41.3 & 93.6 & 80.6 & 92.7 & 78.6 & 90.7 & 65.9
            & \cellcolor[HTML]{d8f3dc}72.2 \\
        \midrule
        \rowcolor{gray!15}
        \multicolumn{14}{l}{\textit{Ours}} \\

        \textbf{POINTS-GUI-G-8B}& \textbf{93.7} & 68.4 & 87.5 & \underline{75.7} & \textbf{85.9} & \textbf{69.9} & \textbf{97.1} & 88.8 & \underline{97.2} & \underline{90.1} & \underline{95.8} & \underline{84.4} & \cellcolor[HTML]{d8f3dc}\underline{87.0}\\
        \bottomrule
    \end{tabularx}
    \vspace{-0.8em}
\end{table*}
\section{Comparison with Other Models}
\begin{table*}[!ht]
    \centering
    \small
    \setlength{\tabcolsep}{2.5pt} 
    \caption{Performance comparison on the \textbf{ScreenSpot-Pro} benchmark. The best results are highlighted in \textbf{bold}, and the second-best results are \underline{underlined}.}
    \label{tab:ssp}

    \begin{tabularx}{\textwidth}{
        l
        *{12}{>{\centering\arraybackslash}X}
        c}
        \toprule
        \multirow{2}{*}{\textbf{Model}} &
        \multicolumn{2}{c}{\textbf{CAD}} &
        \multicolumn{2}{c}{\textbf{Dev.}} &
        \multicolumn{2}{c}{\textbf{Creative}} &
        \multicolumn{2}{c}{\textbf{Scientific}} &
        \multicolumn{2}{c}{\textbf{Office}} &
        \multicolumn{2}{c}{\textbf{OS}} &
        \multirow{2}{*}{\textbf{Avg.}} \\
        \cmidrule(lr){2-3}
        \cmidrule(lr){4-5}
        \cmidrule(lr){6-7}
        \cmidrule(lr){8-9}
        \cmidrule(lr){10-11}
        \cmidrule(lr){12-13}
        & Text & Icon & Text & Icon & Text & Icon & Text & Icon & Text & Icon & Text & Icon & \\
        \midrule
        \rowcolor{gray!15}
        \multicolumn{14}{l}{\textit{Proprietary Models}} \\
        GPT-4o~\citep{gpt4o}
            & 2.0  & 0.0  & 1.3  & 0.0  & 1.0  & 0.0  & 2.1  & 0.0
            & 1.1  & 0.0  & 0.0  & 0.0  & \cellcolor[HTML]{d8f3dc}0.8 \\

        Claude C.~\citep{claude_com}
            & 14.5 & 3.7  & 22.0 & 3.9  & 25.9 & 3.4  & 33.9 & 15.8
            & 30.1 & 16.3 & 11.0 & 4.5  & \cellcolor[HTML]{d8f3dc}17.1 \\

        Gemini-3-Pro~\citep{gemini3pro} 
            & -    & -    & -    & -    & -    & -    & -    & -
            & -    & -    & -    & -    & \cellcolor[HTML]{d8f3dc}\underline{72.7} \\
        Seed1.8~\citep{seed18}& -    & -    & -    & -    & -    & -    & -    & -
            & -    & -    & -    & -    & \cellcolor[HTML]{d8f3dc}\textbf{73.1} \\
        \midrule
        \rowcolor{gray!15}
        \multicolumn{14}{l}{\textit{Open-Source Models}} \\
        Qwen3-VL-2B~\citep{Bai2025Qwen3VLTR} &31.0 &15.6 &55.2 &11.7 &59.1 &16.1 &64.6 &22.7 &72.3 &34.0 &59.8 &23.6 &\cellcolor[HTML]{d8f3dc}41.9 \\
        InfiGUI-3B \citep{infiguig1} & 50.8 & 25.0 & 64.9 & 20.0 & 51.5 & 16.8 & 68.8 & 32.7 & 70.6 & 32.1 & 49.5 & 19.7 & \cellcolor[HTML]{d8f3dc}45.2 \\ 
        Ferret-UI Lite \citep{Ferret-ui-lite}& -    & -    & -    & -    & -    & -    & -    & -
            & -    & -    & -    & -    & \cellcolor[HTML]{d8f3dc}53.3\\

        UI-TARS-7B~\citep{qin2025ui}
            & 20.8 & 9.4  & 58.4 & 12.4 & 50.0 & 9.1  & 63.9 & 31.8
            & 63.3 & 20.8 & 30.8 & 16.9 & \cellcolor[HTML]{d8f3dc}35.7 \\
        Phi-Ground~\citep{phi_ground}
            & 26.9 & 17.2 & 70.8 & 16.7 & 56.6 & 13.3 & 58.0 & 29.1 & 76.4 & 44.0 & 55.1 & 25.8 & \cellcolor[HTML]{d8f3dc}43.2\\
        GUI-Actor-7B~\citep{wu2025gui}
            & 47.7 & 9.4  & 59.1 & 15.9 & 59.6 & 16.1 & 70.1 & 25.5
            & 69.5 & 41.5 & 55.1 & 19.1 & \cellcolor[HTML]{d8f3dc}44.6 \\
        SE-GUI-7B~\citep{SE_GUI}
            & 51.3 & 14.1 & 68.2 & 19.3 & 57.6 & 9.1  & 75.0 & 28.2
            & 78.5 & 43.4 & 49.5 & 25.8
            & \cellcolor[HTML]{d8f3dc}47.2 \\
        GUI-G$^2$-7B~\citep{GUI_G2}
            & 55.8 & 12.5 & 68.8 & 17.2 & 57.1 & 15.4 & 77.1 & 24.5
            & 74.0 & 32.7 & 57.9 & 21.3 & \cellcolor[HTML]{d8f3dc}47.5 \\
        Qwen3-VL-8B~\citep{Bai2025Qwen3VLTR} & 46.7 &10.9 &79.2 &23.4 &68.2 &14.0 &73.6 &30.0 &76.3 &30.2 &65.4 &21.3 &\cellcolor[HTML]{d8f3dc}49.9\\
        OpenCUA-7B~\citep{opencua}
            & -    & -    & -    & -    & -    & -    & -    & -
            & -    & -    & -    & -    & \cellcolor[HTML]{d8f3dc}50.0 \\
        GTA1-7B~\citep{yang2025gta1}
            & 53.3 & 17.2 & 66.9 & 20.7 & 62.6 & 18.9
            & 76.4 & 31.8 & 82.5 & 50.9 & 48.6 & 25.9
            & \cellcolor[HTML]{d8f3dc}50.1 \\

        UI-Venus-7B~\citep{ui_venus}
            & 60.4 & 21.9 & 74.7 & 24.1 & 63.1 & 14.7
            & 76.4 & 31.8 & 75.7 & 41.5 & 49.5 & 22.5
            & \cellcolor[HTML]{d8f3dc}50.8 \\
        InfiGUI-G1-7B~\citep{infiguig1}
            & 57.4 & 23.4 & 74.7 & 24.1 & 64.6 & 18.2 & 80.6 & 31.8
            & 75.7 & 39.6 & 57.0 & 29.2 & \cellcolor[HTML]{d8f3dc}51.9 \\
        GUI-Owl-7B~\citep{ye2025mobile}
            & \underline{64.5} & 21.9 & 76.6 & 31.0 & 59.6 & \underline{27.3} & 79.1 & 37.3 & 77.4 & 39.6 & 59.8 & 33.7 & \cellcolor[HTML]{d8f3dc}54.9 \\
        MAI-UI-2B~\citep{zhou2025maiuitechnicalreportrealworld}&61.4 &23.4 &76.6 &32.4 &69.2 &21.7 &81.2 &34.5 &\underline{85.9} &39.6 &68.2 &41.6 & \cellcolor[HTML]{d8f3dc}57.4\\
        
        MAI-UI-8B~\citep{zhou2025maiuitechnicalreportrealworld}&\textbf{72.6} &\textbf{35.9} &\underline{83.8} &\textbf{52.4} &\textbf{76.3} &\textbf{33.6} &79.9 &37.3 & \textbf{88.7} &\underline{60.4} &\textbf{76.6} & \textbf{49.4} & \cellcolor[HTML]{d8f3dc}65.8\\
        \midrule
        Qwen3-VL-32B~\citep{Bai2025Qwen3VLTR} & 60.4 &\underline{28.1} &69.5 &22.1 &\underline{75.8} &25.2 &\textbf{84.7} &25.5 &\underline{85.9} &43.4 &62.6 &15.7 &\cellcolor[HTML]{d8f3dc}54.9\\

        OpenCUA-32B~\citep{opencua}
            & -    & -    & -    & -    & -    & -    & -    & -
            & -    & -    & -    & -    & \cellcolor[HTML]{d8f3dc}55.3 \\

        GUI-Owl-32B~\citep{ye2025mobile}
            & 62.4 & \underline{28.1} & \textbf{84.4} & 39.3 & 65.2 & 18.2 & \underline{82.6} & \underline{39.1} & 81.4 & 39.6 & 70.1 & 36.0 & \cellcolor[HTML]{d8f3dc}58.0 \\
        UGround-v1-72B~\citep{uground}
            & 16.8 & 4.7  & 55.8 & 4.8  & 54.0 & 10.5 & 70.8 & 22.7
            & 61.0 & 18.9 & 40.2 & 7.9  & \cellcolor[HTML]{d8f3dc}34.5 \\
        UI-Tars-72B~\citep{qin2025ui}
            & 18.8 & 12.5 & 63.0 & 17.2 & 57.0 & 15.4 & 64.6 & 20.9
            & 63.3 & 26.4 & 42.1 & 15.7 & \cellcolor[HTML]{d8f3dc}38.1 \\
        \midrule
        \rowcolor{gray!15}
        \multicolumn{14}{l}{\textit{Ours}} \\
        
        \textbf{POINTS-GUI-G-8B}&46.7 & \underline{28.1} & 76.0 &\underline{44.8} & 66.2 &\textbf{33.6} & 81.3 &\textbf{48.2} & \textbf{88.7} &\textbf{62.3} & \underline{72.0} &\underline{43.8} & \cellcolor[HTML]{d8f3dc}59.9\\
        \bottomrule
    \end{tabularx}
    \vspace{-0.8em}
\end{table*}

In this section, we evaluate \textbf{POINTS-GUI-G} against models of comparable scale across the five benchmarks previously described. All comparative data is sourced from MAI-UI \citep{zhou2025maiuitechnicalreportrealworld}. Overall, POINTS-GUI-G demonstrates robust performance across desktop, mobile, and web environments, either establishing a significant lead over or achieving comparable performance with current state-of-the-art (SOTA) models. Specifically, POINTS-GUI-G ranks first on three of the five benchmarks.

On \textbf{ScreenSpot-Pro} \citep{li2025screenspotpro}, which tests the ability to execute complex tasks within high-resolution interface scenarios, POINTS-GUI-G yields impressive results. It outperforms GTA1-7B \citep{yang2025gta1} by 9.8 points and GUI-Owl-7B \citep{ye2025mobile} by 5 points. Notably, it even surpasses significantly larger models, such as OpenCUA-32B \citep{OpenAICUA}, trailing only MAI-UI-8B on this specific metric.

Regarding complex desktop environments like \textbf{OSWorld-G}\citep{xie2025scalingcomputerusegroundinguser}, POINTS-GUI-G secures the top rank among SOTA models, outperforming MAI-UI-8B by approximately 6 points. Its superiority is also evident in \textbf{UI-Vision}, where it handles diverse and complex instructions with a margin often exceeding 10 points over its competitors. Furthermore, while achieving comparable performance to MAI-UI-8B on \textbf{MMBench-GUI}, POINTS-GUI-G remains the top performer in general grounding tasks that span from mobile to desktop environments (\textbf{ScreenSpotv2}). This consistently high ranking underscores its versatility and effectiveness in GUI grounding across all evaluated scenarios.

\section{Conclusion}
In this work, we introduce POINTS-GUI-G, a high-performance GUI grounding model. Despite its compact 8B-parameter architecture, the model achieves state-of-the-art results, ranking first among similarly sized models across multiple benchmarks and outperforming several significantly larger counterparts. Beyond its performance, we detail our comprehensive optimization methodology, covering data engineering, training strategies, and the specific techniques most critical to enhancing grounding accuracy. To support the research community and address inconsistent evaluation standards, we are open-sourcing our model and evaluation suite. We believe this full-scale contribution will catalyze progress in the field, providing a robust foundation for our future work in optimizing end-to-end task execution for GUI agents.

\clearpage
\newpage
\bibliographystyle{ieeenat_fullname}
\bibliography{paper}

\clearpage
\newpage
\appendix
\section{Appendix}

\subsection{Unified System Prompt for Grounding Task}
\label{sec:prompt}

\definecolor{darkorange}{RGB}{255, 140, 0}
\definecolor{darkblue}{RGB}{84, 112, 198}
\definecolor{lightgreen}{RGB}{145, 204, 117}
\definecolor{lightyellow}{RGB}{250, 200, 88}
\definecolor{lightred}{RGB}{238, 102, 102}
\definecolor{lightblue}{RGB}{115, 192, 222}

\newtcolorbox{promptbox}[2][Prompt]{
colback=black!5!white,
arc=5pt, 
boxrule=0.5pt,
fonttitle=\bfseries,
breakable,
title=#1, 
before upper={\small}, fontupper=\fontfamily{ptm}\selectfont,
colframe=#2, 
}

\begin{promptbox}[Bounding box prediction]{lightgreen}
Output the bounding box in the image of the UI element corresponding to the instruction "Doesn't start checkbox" and the description "A square checkbox with the label 'Doesn't start'" with grounding.\\
\\
Requirements for the output:\\
- Return only the bounding box coordinates (x0, y0, x1, y1)\\
- Coordinates must be normalized to the range [0, 1]\\
- Round each coordinate to three decimal places\\
- Format the output as strictly (x0, y0, x1, y1) without any additional text.
\end{promptbox}

\begin{promptbox}[Center point localization]{lightgreen}
You are a GUI agent. Based on the UI screenshot provided, please locate the exact position of the element that matches the instruction given by the user.\\
\\
Requirements for the output:\\
- Return only the point (x, y) representing the center of the target element\\
- Coordinates must be normalized to the range [0, 1]\\
- Round each coordinate to three decimal places\\
- Format the output as strictly (x, y) without any additional text
\end{promptbox}

\subsection{GUI Grounding Examples}

\begin{figure}[!htbp]
\centering
\includegraphics[width=\linewidth,scale=1.00]{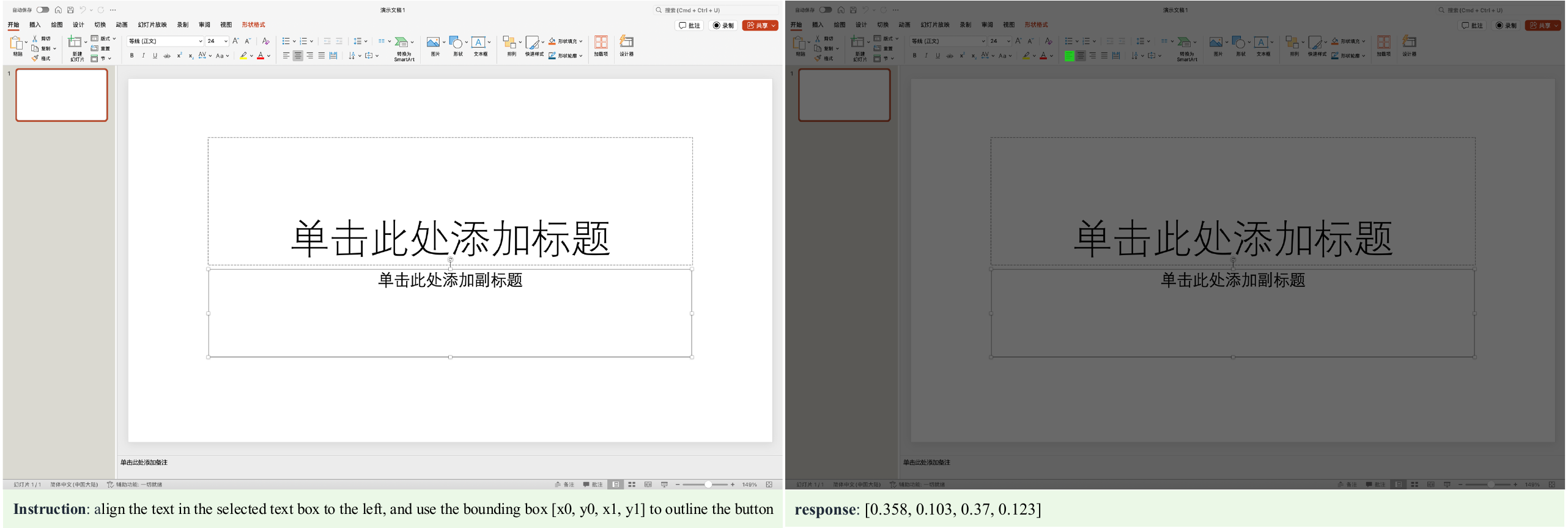}
\includegraphics[width=\linewidth,scale=1.00]{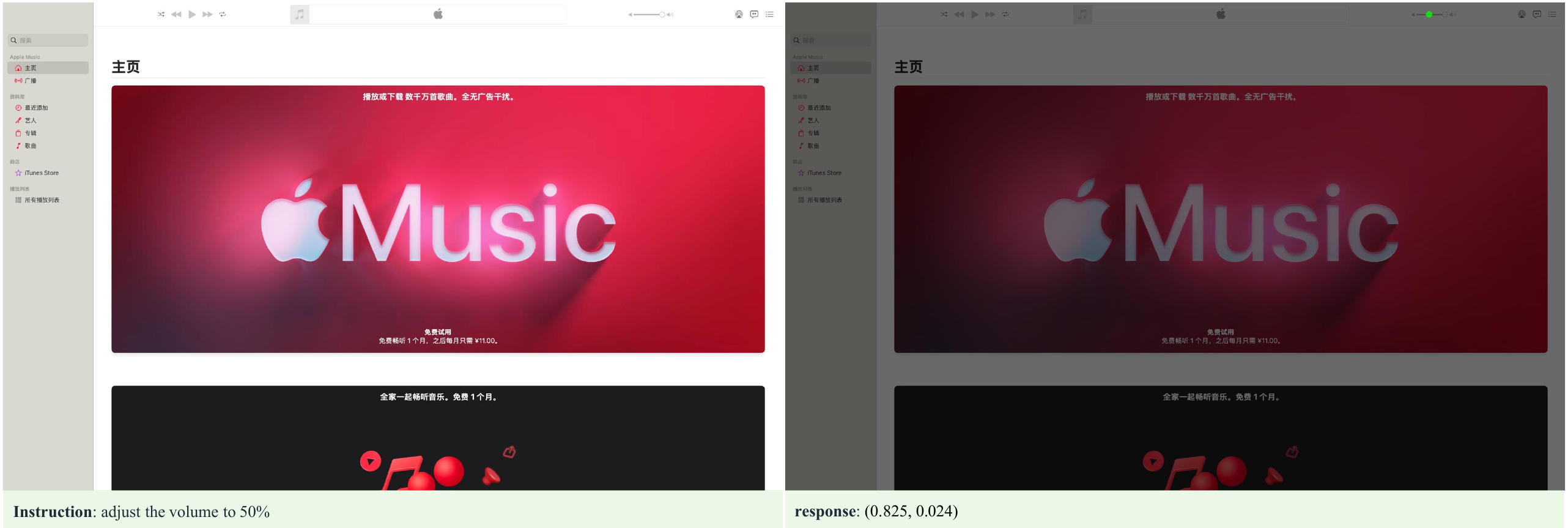}
\includegraphics[width=\linewidth,scale=1.00]{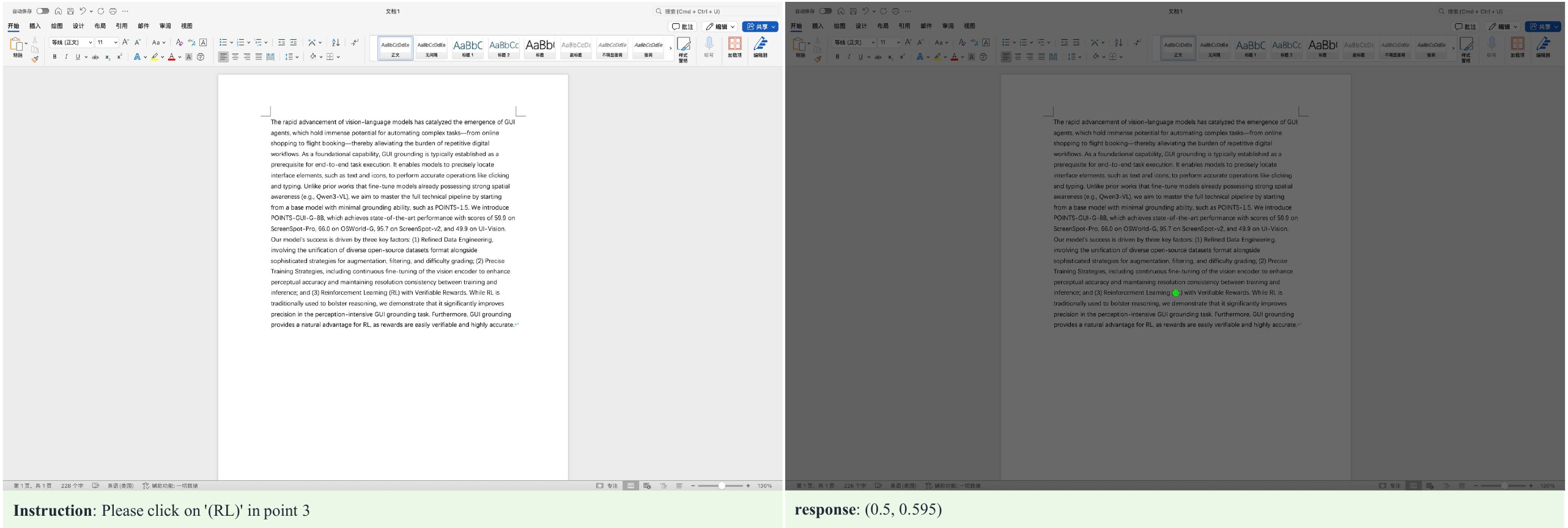}
\captionsetup{justification=centering}
\caption{Prediction on desktop screenshots.}
\label{fig:afig1}
\end{figure}

\begin{figure}[!htbp]
\centering
\includegraphics[width=\linewidth,scale=1.00]{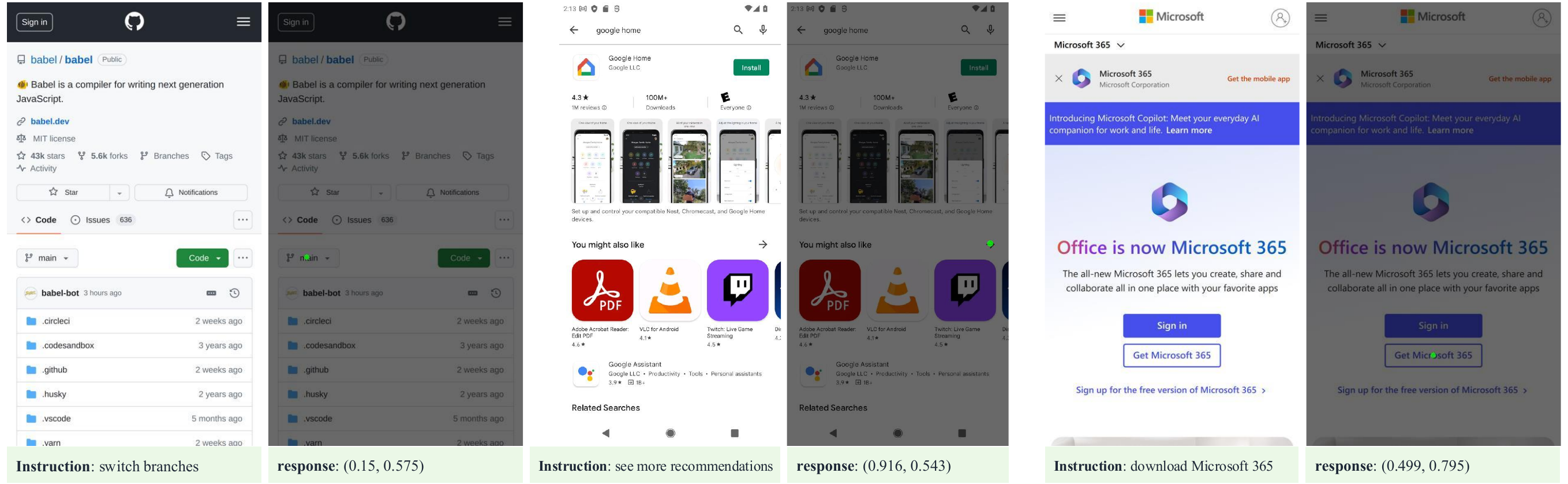}
\captionsetup{justification=centering}
\caption{Prediction on mobile screenshots.}
\label{fig:afig2}
\end{figure}

\begin{figure}[!htbp]
\centering
\includegraphics[width=\linewidth,scale=1.00]{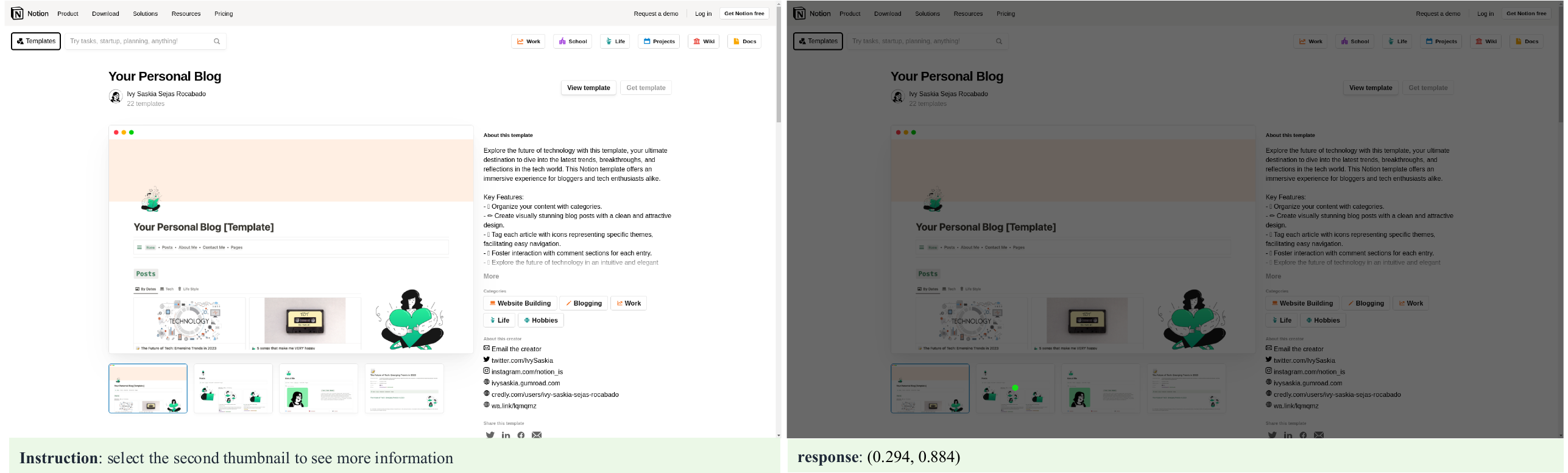}
\includegraphics[width=\linewidth,scale=1.00]{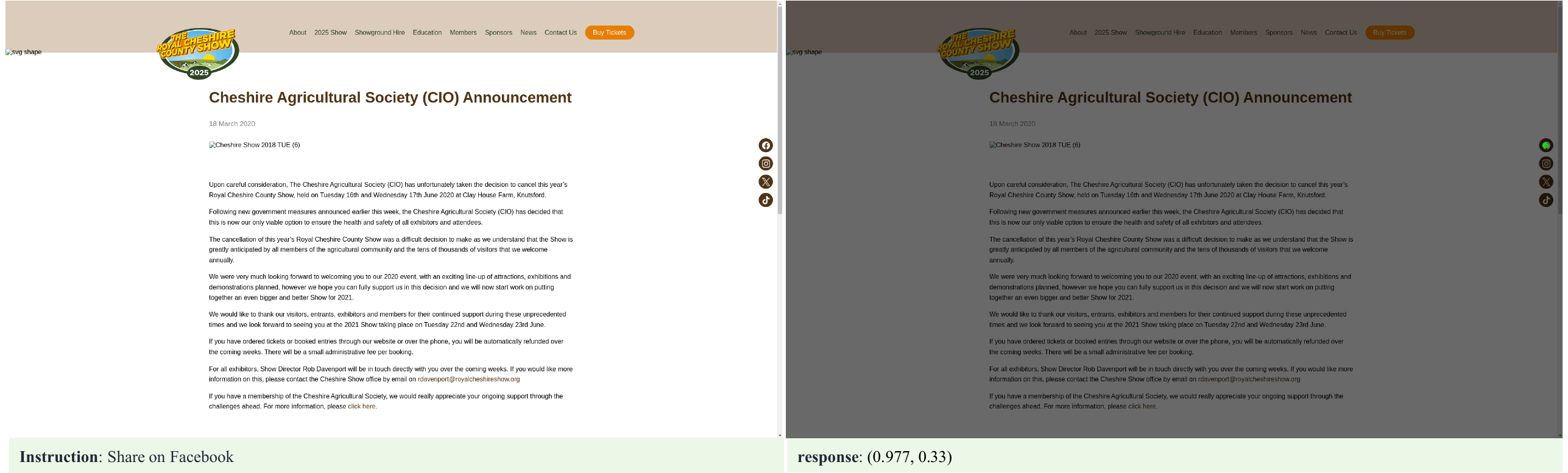}
\includegraphics[width=\linewidth,scale=1.00]{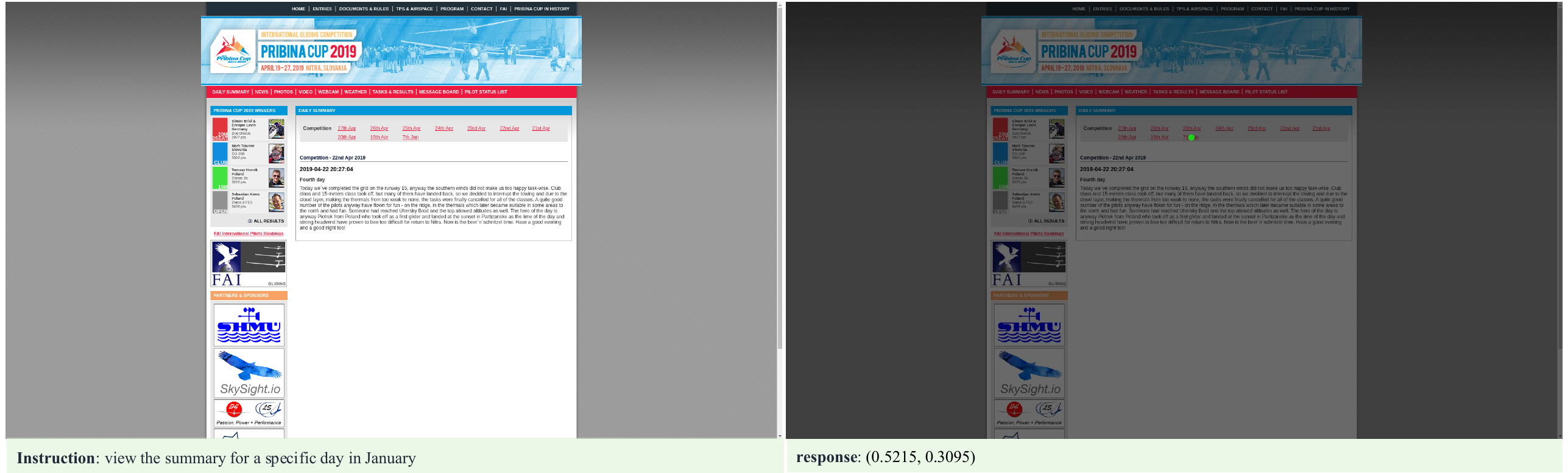}
\captionsetup{justification=centering}
\caption{Prediction on web screenshots.}
\label{fig:afig3}
\end{figure}

\end{document}